\documentclass[remotesensing,article,submit,moreauthors,pdftex]{Definitions/mdpi} 

\firstpage{1} 
\pubvolume{1}
\issuenum{1}
\articlenumber{0}
\pubyear{2025}
\copyrightyear{2025}
\datereceived{} 
\dateaccepted{} 
\datepublished{} 
\hreflink{https://doi.org/} 

\usepackage[utf8]{inputenc}
\usepackage{authblk}
\usepackage{setspace}
\usepackage{amsmath}

\usepackage{caption}
\usepackage{stfloats}
\usepackage{tabularx,booktabs}
\usepackage{array}

\usepackage{subfig}
\newcolumntype{Y}{>{\centering\arraybackslash}X}

\Title{Development and Application of Self-Supervised Machine Learning for Smoke Plume and Active Fire Identification from the FIREX-AQ Datasets}

\TitleCitation{Development and Application of Self-Supervised Machine Learning for Smoke Plume and Active Fire Identification from the FIREX-AQ Datasets}

\Author{Nicholas LaHaye $^{1,2}$\orcidA{}, Anistasija Easley, Kyongsik Yun $^{1}$\orcidB{}, Huikyo Lee$^{1}$\orcidC{}, Erik Linstead$^{3,4}$\orcidI{}, Michael J. Garay $^{1}$\orcidD{}, and Olga V. Kalashnikova$^{1}$\orcidH{}}

\author[1*,2]{Nicholas LaHaye}
\author[2,3]{Anistasija Easley}
\author[2]{Kyongsik Yun}
\author[2]{Hugo Lee}
\author[4,5]{Erik Linstead}
\author[2]{Michael J. Garay}
\author[2]{Olga V. Kalashnikova}

\AuthorNames{Nicholas LaHaye, Anistasija Easley, Kyongsik Yun, Huikyo Lee, Erik Linstead, Michael J. Garay, and Olga V. Kalashnikova }

\AuthorCitation{LaHaye N.; Easley A.; Yun K.; Lee, H.; Linstead, E.; Garay, M. J.; Kalashnikova, O.}

\address{%
$^{1}$ \quad Spatial Informatics Group, LLC., Pleasanton, CA 94566, USA\\
$^{2}$ \quad Jet Propulsion Laboratory, California Institute of Technology, Pasadena, CA 91101, USA\\
$^{3}$ \quad University of California at Berkeley, Berkeley, CA 94720, USA\\
$^{4}$ \quad Fowler School of Engineering, Chapman University, Orange, CA 92866, USA\\
$^{5}$ \quad Machine Learning and Assistive Technology Lab (MLAT), Chapman University, Orange, CA 92866, USA}

\corres{Correspondence:  nlahaye@sig-gis.com}

\abstract{Fire Influence on Regional to Global Environments and Air Quality (FIREX-AQ) was a field campaign aimed at better understanding the impact of wildfires and agricultural fires on air quality and climate. The FIREX-AQ campaign took place in August 2019 and involved two aircraft and multiple coordinated satellite observations. This study applied and evaluated a self-supervised machine learning (ML) method for the active fire and smoke plume identification and tracking in the satellite and sub-orbital remote sensing datasets collected during the campaign. Our unique methodology combines remote sensing observations with different spatial and spectral resolutions. The demonstrated approach successfully differentiates fire pixels and smoke plumes from background imagery, enabling the generation of a per-instrument smoke and fire mask product, as well as smoke and fire masks created from the fusion of selected data from independent instruments. This ML approach has a potential to enhance operational wildfire monitoring systems and improve decision-making in air quality management through fast smoke plume identification and tracking and could improve climate impact studies through fusion data from independent instruments.}

\keyword{FIREX-AQ; fire and smoke detection; Big data applications; clustering; computer vision; image segmentation; self-supervised deep learning; multi-modal data fusion}

\begin{document}


\section{Introduction}

An important and common application of machine learning (ML) is to identify and leverage latent patterns in data or imagery. A typical approach is to use \textit{supervised} learning, which requires a set of truth labels that the ML method attempts to generalize to the problem of mapping from an input dataset $X$ to the output $Y$ through a set of features, $M$. The challenge with supervised learning, and even the recently popularized semi-supervised learning, is acquiring a sufficiently large and \textbf{unambiguous} set of labels, which often requires many hours of manual labor on the part of domain experts. Alternatively, \textit{self-supervised} learning takes a similar input dataset $X$ and finds relationships among the features $M$ resulting in context-free groupings in the output $Y$. Because no labels are provided for the input, there are no labels provided in the output. To utilize the results, the labels or missing context must be assigned after the fact by experts, but this has proven to be a much less labor-intensive endeavor, all while keeping subject matter experts in the loop.

In previous work, we demonstrated that feeding 2-dimensional images of instrument radiances, or Level 1 (L1) data, into Deep Belief Networks (DBNs) coupled with an unsupervised clustering method results in images automatically segmented into relevant geophysical objects \cite{obj_det_init}. We further demonstrated that the same results can be achieved using a simplified architecture across select areas of the globe and for various kinds of land surface and atmospheric segmentation tasks \cite{obj_det_2}. 

In our recent work \cite{encoder_comp}, we have generalized our ML framework into an open-source software system called Segmentation, Instance Tracking, and data Fusion Using multi-SEnsor imagery (SIT-FUSE). This framework allows for various types of encoders, including regular and convolutional DBNs, Transformers, and Convolutional Neural Networks (CNNs), and we have moved from traditional unsupervised clustering to a deep-learning-based clustering approach. 

This approach, as a whole, has several unique benefits. First, it is not restricted to a particular remote sensing instrument with specific spatial or spectral resolution. Second, it has the potential to identify and ``track'' geophysical objects across datasets acquired from multiple instruments. Third, it allows for the joining of data from different instruments, "fusing" the information within the self-supervised encoder. Finally, it can be applied to many different scenes and problem sets, most notably in no- and low-label environments, not just ones for which labeled training sets exist, which is required for strictly supervised ML techniques.

Here we apply our self-supervised ML approach to the problem of automatically detecting and tracking wildfires and smoke plumes, through sequences of L1 (imagery) data acquired by multiple remote sensing instruments during the joint National Aeronautics and Space Administration/National Oceanic and Atmospheric Administration (NASA/NOAA) Fire Influence on Regional to Global Environments and Air Quality (FIREX-AQ) field campaign that took place in the western United States in the summer of 2019 \cite{FIREX_paper}. The high-altitude NASA ER-2 carried seven remote sensing instruments that provided high-spatial resolution observations of fires and smoke plumes in conjunction with NASA DC-8 aircraft and multiple satellite overpasses over same fire events. The FIREX-AQ datasets of collocated satellite and multiple airborne imagery at different spatial resolution are excellent as a testbad for SIT-FUSE method of fire/smoke identification and tracking.

Wildfires and the smoke plumes induced by wildfires substantially contribute to the carbon cycle and can have a long-lasting impact on air quality and Earth's climate system. In addition, human-driven climate change is associated with more frequent and severe wildfires \cite{anthro_fire}. Despite the importance and immediacy of the problem, most research and decision-support tools to study wildfires and plumes use observations from a single instrument whose spatial coverage and (spatial, spectral, and temporal) resolutions vary from very fine to very coarse scales, neither of which, on their own, is fully capable of providing the much-needed information for a comprehensive understanding of wildfires and wildfire smoke \cite{Stavros_fire}. As such, the current study aims to bine datasets with different spatial resolutions from multiple instruments to create a patchwork of datasets that fill the temporal gaps present in current single-instrument fire detection datasets. Here, the first step is testing a general framework for segmenting the datasets from multiple instruments and identifying wildfires and smoke plumes.

The detection and tracking of objects, like wildfires and smoke plumes, within a single instrument data has long required developing instrument-specific retrieval algorithms. Such development is labor-intensive and requires domain-specific parameters and instrument-specific calibration metrics, alongside the manual effort to track retrieved objects across multiple scenes \cite{previous_work_ref}. The recent development of retrieval algorithms is actively underway in the field of supervised deep learning (DL), and various methods (e.g., Convolutional Neural Networks (CNN)s) have been applied. Some of these DL methodologies work well, in terms of precision and accuracy \cite{previous_work_ref}, but are still limited by the requirement that the spatial resolutions between training datasets and output products be the same. These methods also require pre-existing label sets, unlike recent supervised approaches like Fully Convolutional Networks (FCNs), Mask R-CNNs, and Transformers \cite{fcn,mask_rcnn,transformers}, which require large label sets to archive accurate results.

In our previous work, we demonstrated that an encoder trained in a self-supervised manner, namely a Deep Belief Network (DBN), trained with L1 (instrument radiance) images, can segment images based on geophysical objects within the scene, in conjunction with unsupervised clustering \cite{obj_det_init}. The unique benefit of this method is that its application is not limited to a single spatial or spectral resolution, and the method has the potential to detect and track objects from images with different resolutions from multiple instruments. With this method, instead of requiring a per-instrument finely hand-labeled label set, we can apply a coarser manual context assignment after segmentation on a smaller set of training scenes, allowing for this technique to be easily applied in cases of no labels or limited labels. We have also quantitatively validated that the same could be done using a simpler architecture for a set of atmospheric and land surface classification tasks using varying spectral, spatial, temporal, and multi-angle remote sensing data as input \cite{obj_det_2}. Since this work, we have transitioned from unsupervised clustering to self-supervised deep clustering, which we will discuss further in the methods section. This completely self-supervised approach can leverage training data from many different scenes, not just ones that are accounted for by previous label sets for training, as is the case with strictly supervised techniques. Ongoing research applies this self-supervised machine learning methodology to track detected smoke plumes across spatiotemporal domains. However, this study focuses on identifying wildfire and smoke plumes within a single-instrument dataset and using a fusion of datasets from multiple instruments. 

\end{paracol}

\begin{figure}[ht!]
\widefigure
     \subfloat[Map of Fire Location from NASA WorldView Snapshots/MODIS \cite{modis_ref, worldview_ref}]{\includegraphics[width=0.45\textwidth, height=5cm]{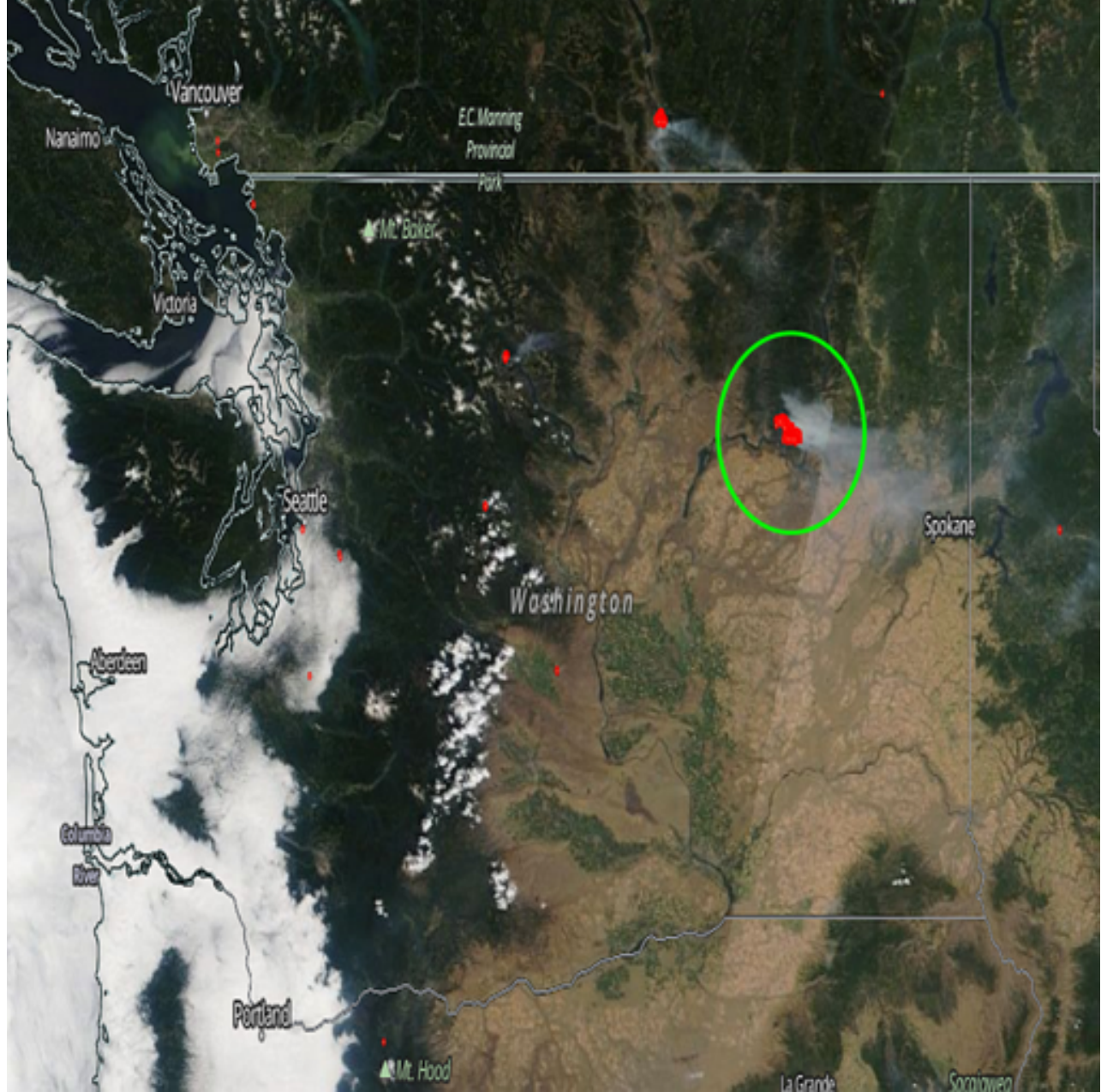}}
     \subfloat[Reference Williams Flat Fire Image from Landsat-8/OLI \cite{ls8_ref}]{\includegraphics[width=0.45\textwidth, height=5cm] {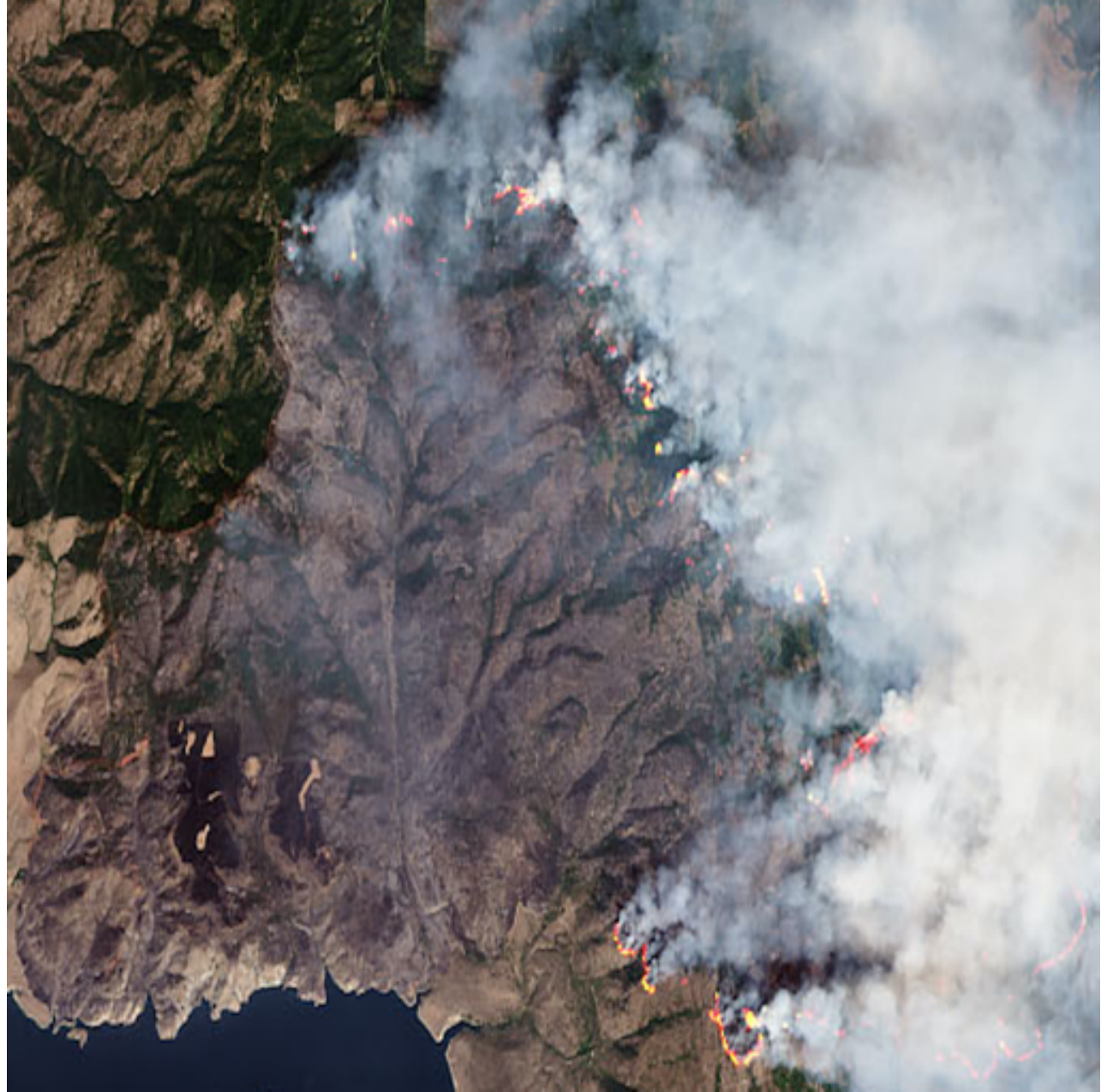}}
     \caption{Reference Map and Imagery}
     \label{fig:1}
\end{figure}
\begin{figure}[ht!]
\widefigure
\centering
     \subfloat{\includegraphics[width=\textwidth, height=4.5cm]{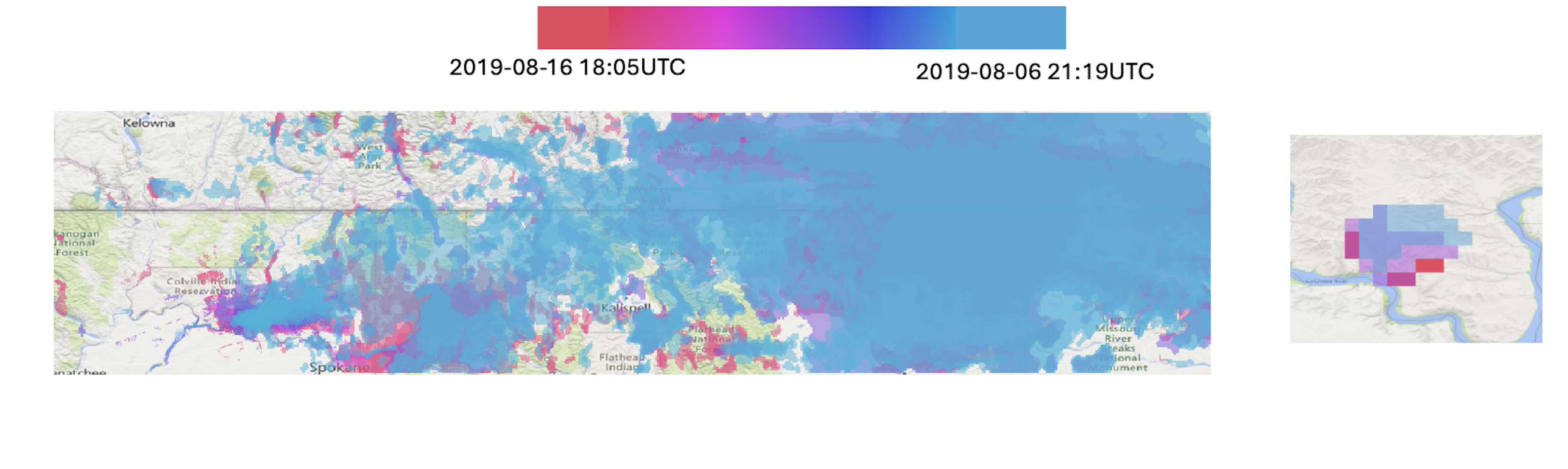}}
     \caption{A display of the smoke plume (left) and the fire front (right) progression of the Williams Flats fire on August 06, 2019, as captured by GOES-17 and segmented via SIT-FUSE.}
     \label{fig:goes_progression}
\end{figure}
\begin{figure}[ht!]
\widefigure
\centering
    \subfloat{\includegraphics[width=0.9\textwidth, height=5cm]{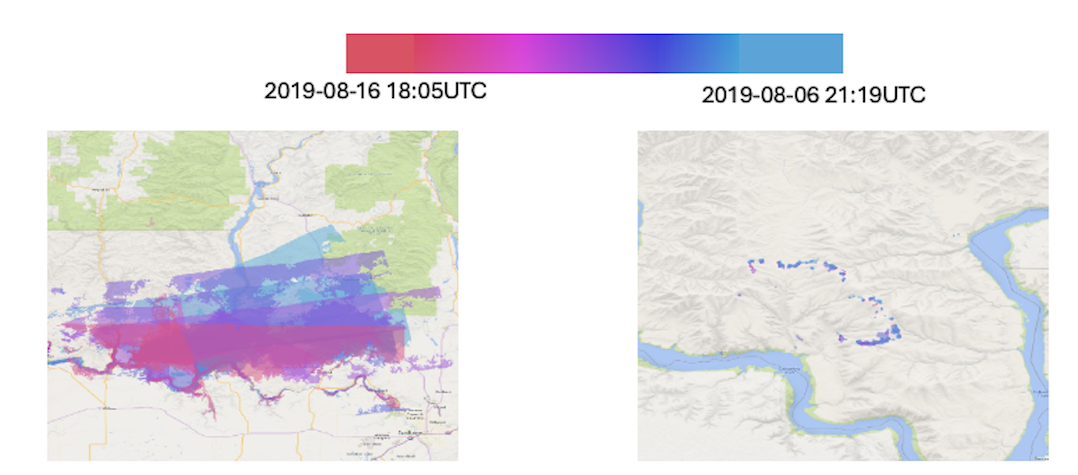}}
    \caption{A display of the smoke plume (left) and the fire front (right) progression of the Williams Flats fire on August 06, 2019, as captured by eMAS and segmented via SIT-FUSE. When used in conjunction with the data in Figure \ref{fig:goes_progression} we can look across large spatial domains at fine temporal scales as well as fine-scale detail at higher spectral resolutions.}
     \label{fig:eMAS_progression}
\end{figure}
\begin{figure}[ht!]
\widefigure
\centering
    \subfloat{\includegraphics[width=0.9\textwidth, height=5cm]{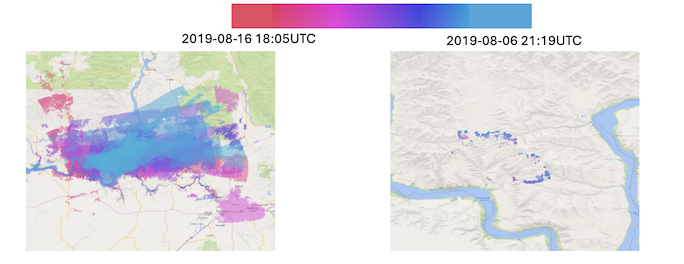}}
    \caption{A display of the smoke plume (left) and the fire front (right) progression of the Williams Flats fire on August 06, 2019, as captured by eMAS, MASTER, AirMSPI, and AVIRIS-C and segmented via SIT-FUSE. This combination of instrumentation maximizes the temporal resolution at the associated high spatial and spectral (/polarization) resolutions of the airborne instrumentation. When compared to the data in Figure \ref{fig:eMAS_progression}, this increases the yields even more for monitoring and science capabilities when used in conjunction with geospatial and polar orbiting instrumentation.}
     \label{fig:multi_sensor_progression}
\end{figure}

\begin{paracol}{2}
\nolinenumbers
\switchcolumn

We used remote sensing datasets from airborne and satellite instruments collected during the Fire Influence on Regional to Global Environments Experiment - Air Quality 2019 (FIREX-AQ 2019; \url{https://csl.noaa.gov/projects/firex-aq/}) campaign. Specifically, we investigated the wildfires and smoke plumes originating from the Williams Flat Fire over the three days from August 6th through August 8th, 2020. Figure \ref{fig:1} shows a map of the active fire (red area within the green circle), taken from NASA's WorldView Snapshots web tool and a close-up reference image of Williams the fire taken from the Landsat-8 Operational Land Imager (OLI). Where available, we also tested on scenes over the Sheridan Fire, from August 15 through August 23. The FIREX-AQ campaign involved two aircraft: the DC-8 aircraft with a primary payload of in-situ instruments supplemented by several remote sensors, and NASA's high-altitude ER-2 aircraft with a package of seven remote sensing instruments. The ER-2 aircraft coordinated on a few occasions with the DC-8 aircraft and had multiple collocations with satellite sensors flying directly along the satellite track. Therefore, several observations of the same active fires and smoke plumes were made by multiple instruments at various spatial resolutions. Table \ref{tab:1} and Table \ref{tab:2} summarize the datasets from airborne and satellite instruments used in this study. 
\noindent
\belowrulesep = 0mm
\aboverulesep = 0mm
\begin{table}[bt!]
 \caption{Airborne instruments and their products.}
 \centering
\begin{tabularx}{\linewidth}{YYYY}
\hline
\textbf{Platform} & \textbf{Instruments} &\textbf{Science Products} & \textbf{Spatial Resolution} \\
\hline
NASA ER-2 &\raggedright Airborne Multiangle SpectroPolarimetric Imager (AirMSPI) \cite{MSPI_ref}&Spectro-polarimetric intensities (10 m spatial resolution, 8 wavelengths in 355-935 nm spectral range, 3 polarimetric bands)& 10m \\
NASA ER-2 &\raggedright Enhanced MODIS Airborne Simulator (eMAS) \cite{eMAS_ref}&Spectral intensities in 38 bands in 445nm - 967nm and 1.616$\mu$m-14.062$\mu$m spectral ranges&50m \\
\hline
NASA DC-8&\raggedright MODIS/ASTER Airborne Simulator (MASTER) \cite{MASTER_ref}&Spectral intensities in 50 bands in 0.44-12.6$\mu$m spectral range &10-30m \\
NASA DC-8&\raggedright  Airborne Visible / Infrared Imaging Spectrometer - Classic (AVIRIS-C) \cite{AVIRIS_ref}&Spectral intensities in 224 bands in 400-2500 nm spectral range &10-30m \\
\hline
\label{tab:1}
\end{tabularx}
\end{table}

\noindent
\belowrulesep = 0mm
\aboverulesep = 0mm
\begin{table}[bt!]
 \caption{Satellite instruments and their products}
 \centering
\begin{tabularx}{\linewidth}{YYYY}
\hline
\textbf{Platform} & \textbf{Instruments} &\textbf{Science Products} & \textbf{Spatial Resolution} \\
\hline
Terra & Multi-angle Imaging SpectroRadiometer (MISR) \cite{MISR_ref}& Spectral intensities in 446nm, 558nm, 672nm, and 867nm& 1.1km and 275m, all resampled to 1.1km \\
\hline
Terra and Aqua& MODerate resolution Imaging SpectroRadiometer (MODIS) \cite{modis_ref}&Spectral intensities in 38 bands in 445nm-967nm and 1.616$\mu$m-14.062$\mu$m spectral range&1km when used alone and resampled to 1.1km when used with MISR \\
\hline
Suomi NPP and NOAA-20 & Visible Infrared Imaging Suite (VIIRS) \cite{VIIRS_ref}&Spectral intensities in 5 bands in 0.44-12.6$\mu$m spectral range &375m \\
\hline
GOES-17& Advanced Baseline Imager (ABI) \cite{GOES_ref}&Spectral intensities in 50 bands in 0.47-13.3$\mu$m spectral range &1km and 2km, all resampled to 2km \\
\hline
PlanetScope& Dove Imagers \cite{planet_ref}&Spectral intensities in 4 bands in 455-860 nm spectral range&1-3m native resolution, all resampled to 10m per the NASA Technical Report \hreflink{https://ntrs.nasa.gov/citations/20240001694}{Planet Imagery Geometric Assessment}\\
\label{tab:2}
\end{tabularx}
\end{table}

This approach not only allows us to leverage single- and multi-instrument datasets to create a denser static patchwork of fire and smoke detections with increased spatial, spectral, and temporal resolution (as depicted in Figures \ref{fig:goes_progression} \ref{fig:eMAS_progression} \ref{fig:multi_sensor_progression}), but it also gives us a uniform embedding-based representation of the data via the encoder outputs and final output of clusters. The final cluster output can be used in conjunction with spatial distributions of the output labels to facilitate fire and smoke plume instance tracking across multi-sensor scenes over varying spatiotemporal domains. Figures \ref{fig:goes_progression} \ref{fig:eMAS_progression} \ref{fig:multi_sensor_progression} demonstrate the various tiers and scales of representative capabilities over the Williams Flats fire on August 06, 2019, when incorporating observations GOES at the coarse spatial but fine temporal resolution end of the scale, and the airborne instruments mentioned in Table \ref{tab:1} over the Williams Flats and Sheridan fires at the fine spatial but coarse temporal end of the scale, along with the polar orbiters in-between these two extremes.

\end{paracol}

\begin{figure}[ht!]
\widefigure
\centering
    \subfloat{\includegraphics[width=0.9\textwidth, height=5cm]{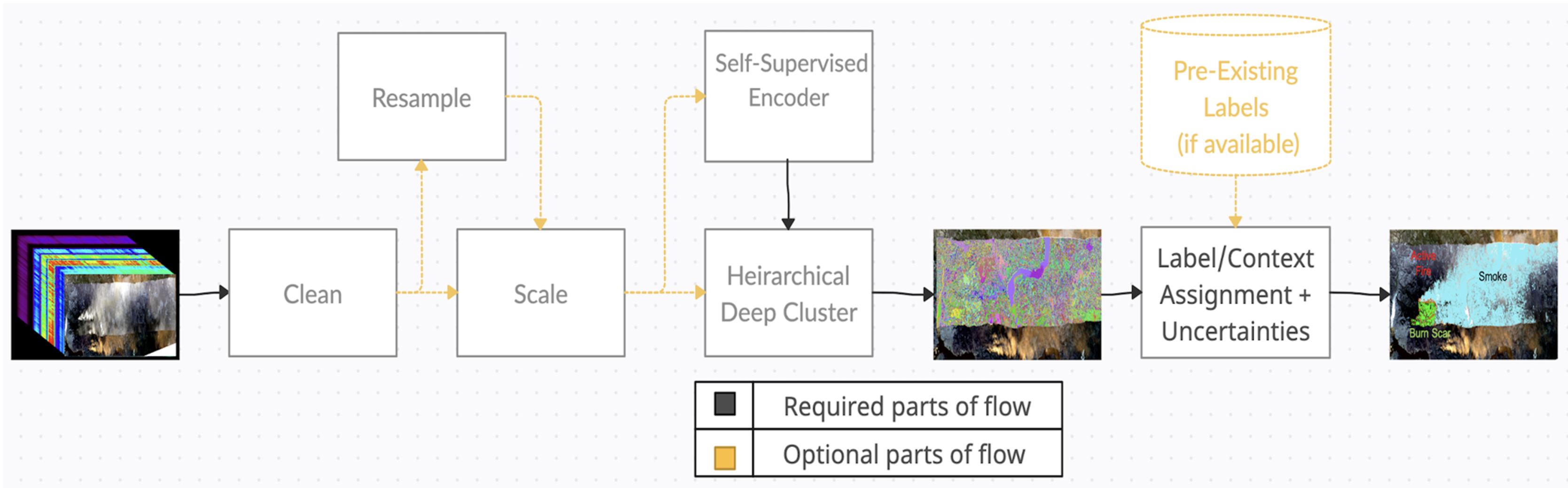}}
    \caption{A flow diagram for the processing of one input type (single instrument or fusion set) through SIT-FUSE}
     \label{fig:flow}
\end{figure}

\begin{paracol}{2}
\nolinenumbers
\switchcolumn

\section{Related Works}

Work on the general problem of self-supervised image segmentation appears to have had success in separating the foreground from the background \cite{segmentation_1, segmentation_2}, or have only used single bands of input from one type of instrumentation, which is effective for their applications, but does not cover the breadth required here \cite{segmentation_3}. Other works have aimed to perform tasks, such as outlining buildings and roadways \cite{segmentation_4}, which is not the goal here. A similar study, which used just autoencoders and a form of clustering, an overall architecture that is close to a subset of ours, attained an accuracy of ~83\% on Landsat imagery alone \cite{segmentation_5}. This uses a similar kind of model to our studies, but uses a single instrument. With large variations in spatial and spectral resolutions, our technique attains higher accuracy (and balanced accuracy, in some cases) across many different instrument sets, including fused data. Even with more recent breakthroughs in semi-supervised semantic segmentation, like the Segment Anything Model (SAM), a problem-dependent amount of labels is required, and SAM is largely unproven in complex domains like remote sensing \cite{SAM}. The identification of the necessary size of label sets, generation of per-pixel label sets, and testing of the feasibility of new techniques in more complex domains are all problem-specific and time-consuming tasks that can be skipped, given our solution - as seen in the successful but extremely limited cases discusses in \cite{CNN_hyper_1, CNN_hyper_2, uav_fire, google_fire}. Lastly, there are new physics-based retrieval techniques, which seem promising, but need continued rigorous analysis to generalize across different regions and instrument types \cite{smoke_phys}. In the future, it may be useful to combine the physical parameterizations and ML-based retrievals via ML loss functions that are 'physics-aware'. The lack of need for large new label sets mitigates the costly, labor-intensive work of manually segmenting each pixel within a dataset used for ground truth, a process which is itself error-prone, and other previously mentioned supervised learning-related precursors model training. Also, leveraging pre-existing operational products to use as labels for supervised learning tasks will inherently cause them to either lack training set diversity or suffer from the issues mentioned above. On the other hand, our approach is well suited to handle large amounts of data, because our unsupervised and self-supervised models can perform label-free image segmentation. The fact that the human-in-the-loop steps of context application and validation occur after the images have been segmented allows for human oversight while mitigating the need for the extremely labor-intensive act of pixel-by-pixel manual segmentation for tens of thousands of images.

\section{Materials and Methods}

\subsection{Methods}

\subsection{Data Preprocessing}
For single instrument cases, data was re-projected to the WGS84 Latitude/Longitude projection. For multi-sensor cases, data is collocated, re-projected to the WGS84 Latitude/Longitude projection, resampled to the lowest common spatial resolution, and stacked channel-wise. The actual fusion occurs as a part of the representation learning done inside the encoder. 

Samples were generated by taking pixels and their direct neighbors (center pixel + 8 neighbors), in all channels, and creating a flat vector. These samples were then standardized by removing the per-channel mean and scaling to per-channel unit variance. The per-channel mean and standard deviations are computed over the full sample set from the scenes used for training. 

For training, the scenes used are subsets. To ensure the subset of samples from the training scenes contained a representative variety of terrain and phenomena, k-means clustering was applied to the data with a set of 50 classes. This stratification technique was chosen as it has proven to be an effective, albeit naive, way to ensure variation within training samples \cite{MacQueen1967, Lin0HG21, 2023AGUFMIN51C0429B}. Three million samples are randomly sampled based on the stratification generated by the full number of samples from the training scenes labeled with the 50 k-means classes. All pixels that are set to fill values, or out of specified valid ranges are discarded before any pre-processing. Regarding the spectral bands used, all spectral bands were used except bands that were extremely noisy or known to be non-functional for the FIREX-AQ campaign time frame. The same pixels used to train the encoders are also used to train the deep clustering heads, and the full training scenes are used to assign context. 

SIT-FUSE can take larger tiles for Convolutional DBNs, CNN-based, and Transformer-based architectures (as mentioned below). However, the pixel neighborhood has proven to provide enough spatial context with regular DBNs, as used in this work. Figure \ref{fig:flow} depicts the overall flow of SIT-FUSE.

\subsubsection{Self-Supervised Representation Learning}
SIT-FUSE is developed to be a generic framework allowing various kinds of encoders and foundation models that leverage self-supervised representation learning, including Deep Belief Networks (DBNs) trained using contrastive divergence, Convolutional Neural Networks (CNNs) with residual connections trained via Bootstrap Your Own Latent (BYOL), and Transformers trained using Image-Joint Embedding Predictive Architecture (I-JEPA) or Masked AutoEncoders (MAEs), as well as pre-trained Transformers for Earth Science, like \hreflink{https://madewithclay.org/}{Clay} \cite{grill2020bootstrap, abs-2301-08243, liao2022gaussianbernoullirbmstears, HeCXLDG22}. For all of these experiments, we used DBNs with 2-3 layers. DBNs were selected here because previous work and experiments done in this work demonstrated that they produce reasonable results and the parameter space is much smaller than the other models mentioned above. We have done extensive validation on the use of DBNs from both the perspective of structural understanding and downstream task performance, as well as resource consumption assessment, for the large set of single-instrument and fusion datasets \cite{obj_det_2}. In short, 2-3 layer DBNs provide a relatively compact model (~2 million parameters vs ~100 million - 10 billion parameters) with representational capabilities that meet our needs, and have demonstrated out-sized representational capabilities in other experiments relative to much larger models \cite{liao2022gaussianbernoullirbmstears}. We are currently evaluating encoder complexity in relation to segmentation performance and geographic coverage, to optimally operationalize this approach for operational global production, which will be discussed further in later manuscripts.

As in our previous work \cite{obj_det_2, obj_det_init, encoder_comp}, the DBN architectures used leverage feature expansion, outputting embeddings of a pixel neighborhood in a larger feature space than the one with which the samples were input. Similar to the idea behind input kernelization, architecture-based feature expansion allows the models to learn nonlinear latent patterns that would be compact and complex in lower dimensions in simpler but higher-dimension forms. This method, although not the most common use case for DBNs or encoders in general, has been demonstrated effectively in other studies as well as in our previous work \cite{feature_exp1, feature_exp2, feature_exp3, obj_det_init, obj_det_2}. In previous works, we held architecture parameters static for all tests to demonstrate efficacy. Here, we varied the number of layers and hidden/output parameters for each encoder used, based on input data spectral resolution. 

\end{paracol}

\begin{paracol}{2}
\nolinenumbers
\switchcolumn

\subsubsection{Deep Clustering}
To extract segmentation maps from the per-pixel embeddings, we use deep clustering, specifically Information Invariant Clustering.

Previous experiments used BIRCH and other forms of traditional agglomerative clustering. We have transitioned to DL-based clustering because the training time, inference time, memory requirements, and model re-usability are all much improved when using neural network layers via PyTorch when compared to using traditional clustering via sci-kit-learn. These layers are trained using the invariant information clustering (IIC) loss function. IIC aims to assign labels that maximize mutual information between an input sample x and a perturbed version of x, x’ \cite{JiVH19} . For our use, perturbations are additions of Gaussian noise to the outputs of RBM-based architectures.

\end{paracol}
\begin{figure}[ht!]
\widefigure
\centering
    \subfloat{\includegraphics[width=0.6\textwidth, height=5cm]{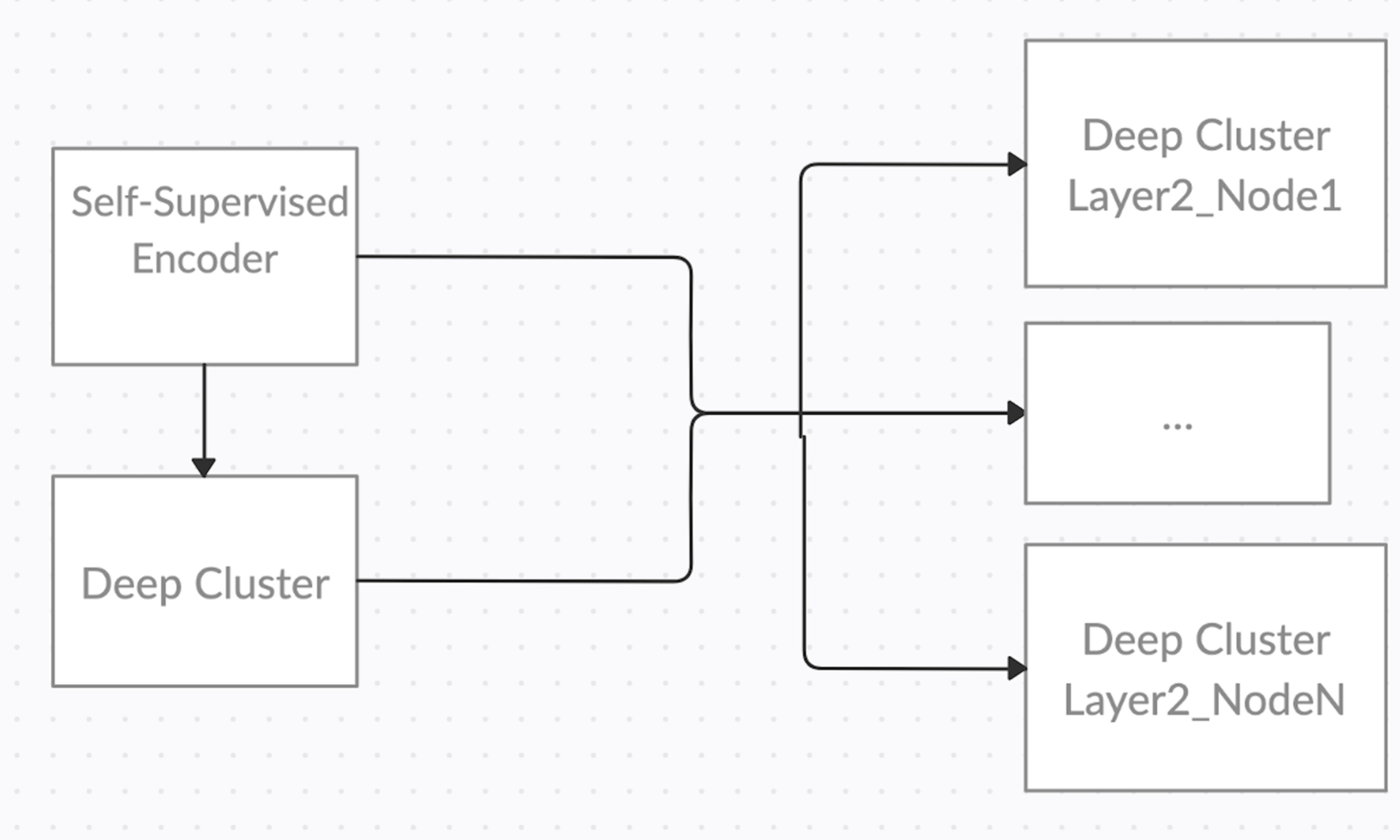}}
    \caption{A 2-layer example of the setup for hierarchical deep clustering. Each box labeled ‘Cluster’ is a set of fully connected layers, connected to the RBM-based model and trained via the IIC loss function. Each child node is only trained and makes predictions on samples given the label from its parent nodes that are associated with the path of edges that link the root to the current node. This setup allows us to use deep clustering to create interlaced levels of specificity for data exploration and characterization}
     \label{fig:iic_heir}
\end{figure}
\begin{paracol}{2}
\nolinenumbers
\switchcolumn

To mimic the hierarchical nature of agglomerative clustering output, we have set up hierarchical deep clustering layers. Here, the output heads are set up in a tree structure where each sub-tree is only trained on label samples classified as belonging to their parent label sets. Each layer receives only the output of the encoder, but as the tree is built from the top down, each neural network in a child node position on the tree only sees samples assigned labels associated with its parent and ancestor node(s). In this way, we can create scene segmentation at varying levels of specificity and explore the connections between each level. To our knowledge, this is the first study/ software system that leverages IIC layers, or even deep clustering layers, in this fashion. For each of the pipelines, the root IIC head allows for 800 possible classes and each child IIC head allows for 100 subclasses. Currently, the level of hierarchy is specified by the user and there is no automated node splitting, but this could be implemented in the future. For our purposes, only two levels of hierarchy are used. We believe that this not only allows us to attain the specificity required for our segmentation tasks but that the hierarchical labeling can be leveraged as a co-pilot for data exploration and discovery. Figure \ref{fig:iic_heir} shows a two-level tree for performing deep clustering.
\subsubsection{Hand Labeling}

For both context assignment and validation, subject matter experts labeled areas of high-certainty smoke, fire, and associated backgrounds. All areas that labelers were uncertain about, remained without labels. This labeling process was done by generating polygons over the remote sensing imagery. An example is provided below in Figure \ref{fig:labeling}. Because scenes can have overlapping classes (i.e. fire and smoke are contained in the same pixel), but also have areas distinct to a single class, a separate background class label set was generated for fire and smoke. Figure \ref{fig:labeling} illustrates the labeling process over a single AVIRIS-C scene.

\end{paracol}

\begin{figure}[ht!]
\widefigure
\centering
    \subfloat{\includegraphics[width=0.7\textwidth, height=5cm]{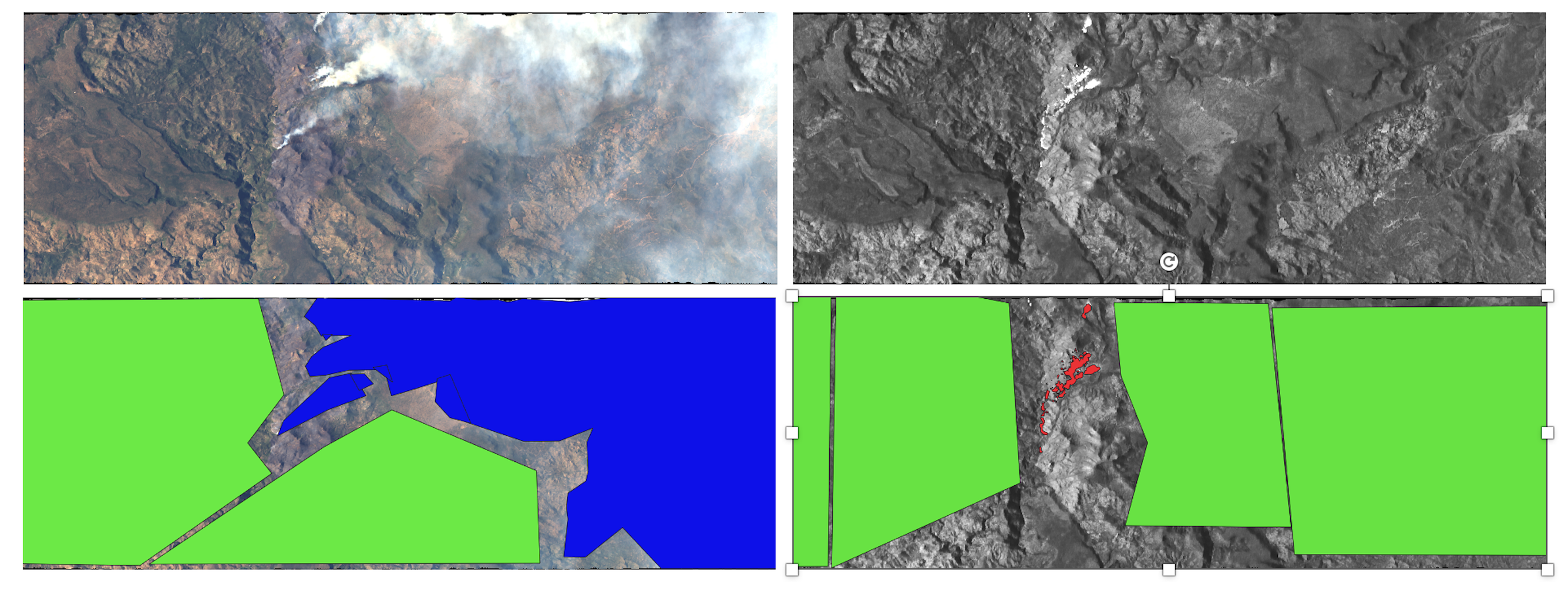}}
    \caption{Visualization of the labeling process. The top row depicts a reference RGB generated from an AVIRIS-C scene (top left) and a single thermal band used for identifying fire front locations (top right). The corresponding image in the bottom row has the smoke and smoke background labels over the RGB image and the fire and fire background labels over the thermal image.}
     \label{fig:labeling}
\end{figure}

\begin{figure}[ht!]
\widefigure
\centering
    \subfloat{\includegraphics[width=0.7\textwidth, height=5cm]{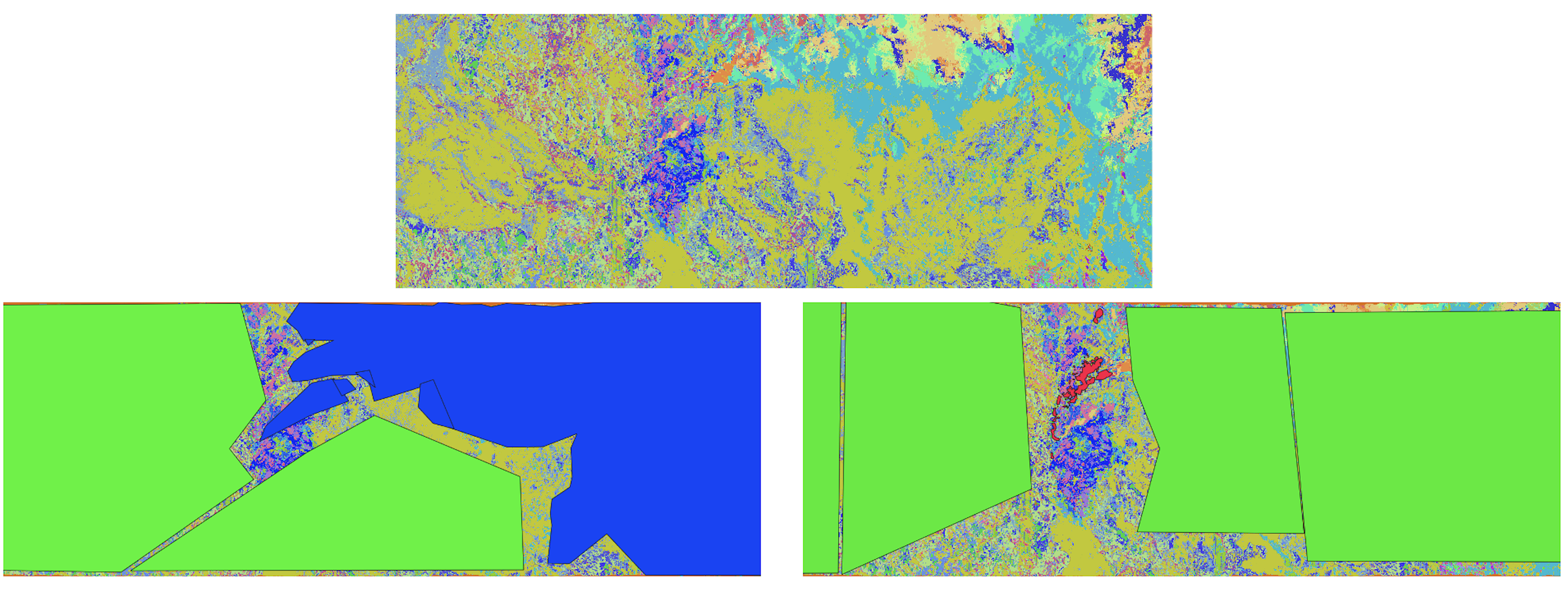}}
    \caption{Visualization of the overlay required for the zonal histogramming process. The top row contains the context-free segmentation map generated via SIT-FUSE for the scene in Figure \ref{fig:labeling}. The bottom row contains the same map with the smoke/background (left) and fire/background (right) labels overlayed. The zonal histogramming process provides counts of classes in the segmentation map relative to label positions and each class is assigned to the label it best agrees with over multiple scenes.}
     \label{fig:context_assign}
\end{figure}

\begin{paracol}{2}
\nolinenumbers
\switchcolumn

Unlike labeling for supervised learning, this approach does not require all training samples to be labeled, which is relevant for problems with high uncertainty in boundary cases, like the segmentation of fire fronts and smoke plumes. The labeling of only the high certainty class areas allows us to capture and compare against segmentation structure, and provide ample samples for context assignment. Although this labeling minimizes the pre-analysis labor required from subject matter experts, it still keeps experts in the loop (a crucial piece for science-related ML tasks). Relative to the number of scenes labeled here, supervised and semi-supervised tasks require 100x more labeled samples or more. Also, because they are learning the mapping between the labels and the input datasets directly, they require much more complete label sets (i.e. uncertain areas must be labeled background or foreground, potentially leading to systematic over-segmentation or under-segmentation).

\subsubsection{Context Assignment}

To assign context to the context-free segmentation maps generated via SIT-FUSE, we used zonal histogramming. For this, we overlayed the labels for a specific scene on the segmentation map generated for the same scene and generated counts of overlapping occurrences of the segmentation map classes and the high-certainty smoke, fire, and background labels. These were gathered over multiple scenes, and then each class in the segmentation maps was assigned to the labels it best matched. This assignment was done separately for each class of interest and its associated background class to ensure multi-class representation, like pixels containing both smoke and fire. Figure \ref{fig:context_assign} provides an example of labels overlayed on a segmentation map.

\end{paracol}

\begin{figure}[ht!]
\widefigure
\centering
     \subfloat{\includegraphics[width=0.85\textwidth, height=8cm]{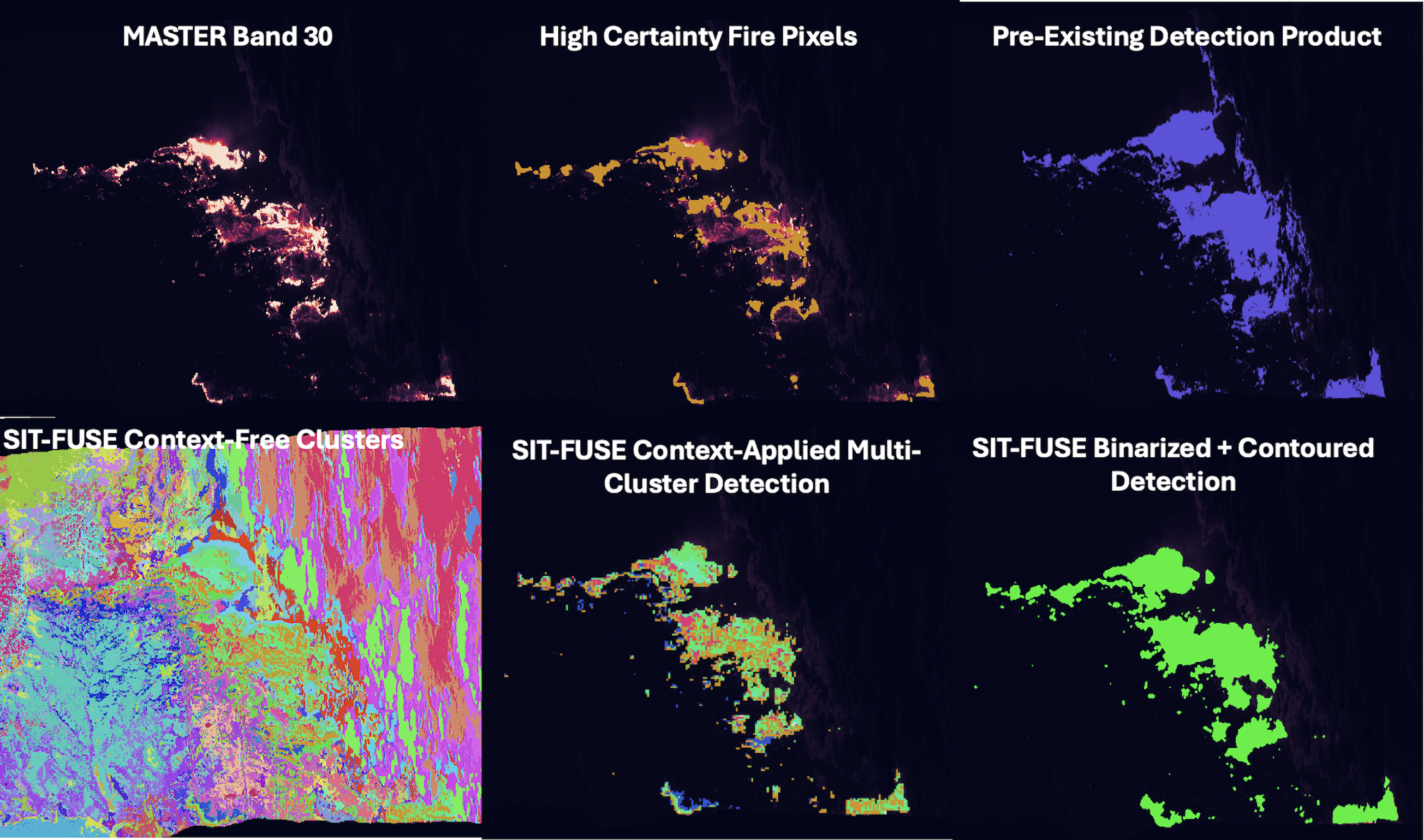}}
     \caption{The top left panel is a single pseudo-colored thermal band from MASTER from a scene over the Williams Flats fire. Following this panel, on the top row are the hand-labeled high-certainty fire pixels overlayed on the MASTER band, and the pre-existing band-ratio-based detection overlayed on the MASTER band, respectively. The bottom row consists of the context-free segmentation map generated from SIT-FUSE for this scene, the context-applied multi-cluster detection - a subset of the full segmentation map with only the labels/clusters that correlate to fire fronts, and the final binarized and contoured fire mask.}
     \label{fig:master_wf}
\end{figure}

\begin{figure}[ht!]
\widefigure
\centering
    \subfloat{\includegraphics[width=0.85\textwidth, height=8cm]{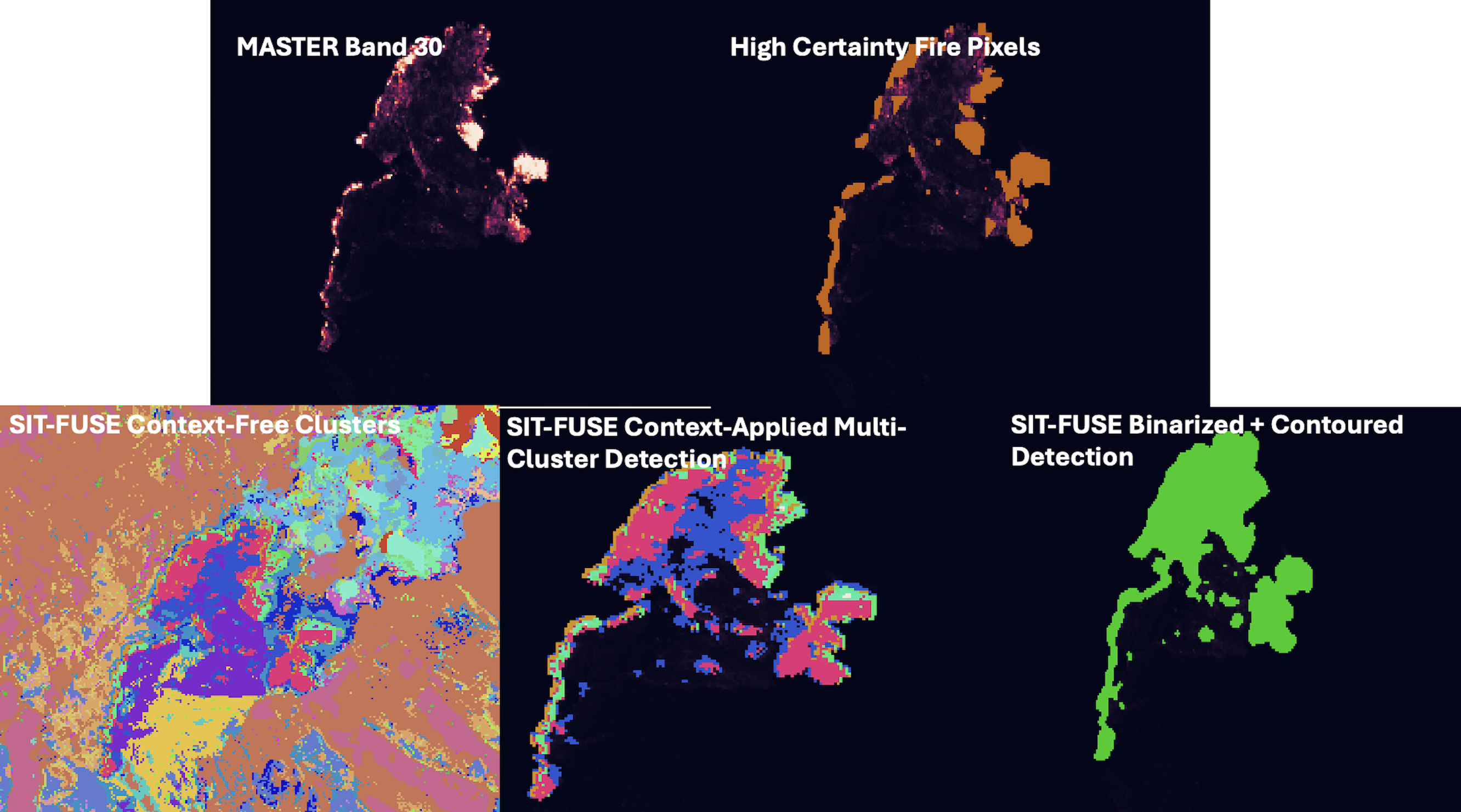}}
     \caption{The top left panel is a single pseudo-colored thermal band from MASTER from a scene over the Sheridan fire. Following this panel, on the top row are the hand-labeled high-certainty fire pixels overlayed on the MASTER band, and the pre-existing band-ratio-based detection overlayed on the MASTER band respectively. The bottom row consists of the context-free segmentation map generated from SIT-FUSE for this scene, the context-applied multi-cluster detection - a subset of the full segmentation map with only the labels/clusters that correlate to fire fronts, and the final binarized and contoured fire mask.}
     \label{fig:master_sheridan}
\end{figure}

\begin{paracol}{2}
\nolinenumbers
\switchcolumn

\subsubsection{Contouring and Filtering}
As the pixels within the detection are components of larger, often connected, geophysical objects, SIT-FUSE allows for building objects from separate detection pixels. To do this, we leverage the contouring capabilities within the openCV python / C++ software package \cite{SUZUKI198532, opencv_library}. Once generated, these contours are filled and we are left with multiple separate objects, instead of many more distinct pixels. We can also filter out small and large objects, where appropriate, and while not used here, leveraging dilations and erosions on the contoured objects is also possible. To mimic the automated process of an operationalized version and ensure fair comparisons and validation, all processing is performed uniformly across all scenes in an input set.

\subsubsection{Validation}
There is no direct ground truth here - meaning comparisons are done against pre-existing detections as well as hand labels generated over only the high certainty areas of smoke plumes and fire fronts - and we know there are other areas of these objects that are not included in the labels. Therefore, recall, precision, and their collective summary via F1-scoring are too harsh of evaluators here. We evaluated some previously published versions of precision and recall that apply fuzzy logic, but ultimately landed on the structural similarity index (SSIM) to evaluate performance across the various dataset pairs\cite{Wang2004,4775883}. This is a fairly common problem within the remote sensing domain and one we aim to help solve with the collective incorporation of self-supervised learning, subject matter expert domain knowledge, and large amounts of data \cite{JRS.15.041501, yang2019evaluatingexplanationgroundtruth}.

\subsection{Materials and Tools}
The software was developed with Python 3.9.13. SIT-FUSE has open-source functionality at its core \cite{sit_fuse_software}. To achieve the required goals of the software and leverage pre-existing and well-validated open-source software, geospatial, big data, and ML toolkits are the backbone of SIT-FUSE. For optimized handling and computation on large datasets across CPUs and GPUs, numpy, scipy, dask, xarray, Zarr, numba, and cupy are used \cite{numpy, scipy, Dask, xarray, zarr, numba, cupy} . For CPU- and GPU-based ML model training, deployment, evaluation, and auto-differentiation, sci-kit-learn, PyTorch, and torchvision are used \cite{pytorch, sklearn}. Because RBMs are not included within the PyTorch library, Learnergy, an open-source library that contains various PyTorch-backed RBM-based architectures is used as well \cite{learnergy}. On the geospatial side of the problems being solved, pyresample, GDAL, OSR, healpy, polar2grid, and GeoPandas are leveraged \cite{pyresample, gdal, healpy, polar2grid, geopandas}. Lastly, for non-machine-learning computer vision techniques, OpenCV is used \cite{opencv}. The combination of these commonly used and well-tested software systems allows us to employ state-of-the-art approaches and architectures with minimal development and maintenance efforts, most of which are only minimally visible to the end user. SIT-FUSE is also publicly available and maintained on the public version of GitHub. For labeling, context assignment, and visualization / qualitative assessments, QGIS, an open-source Geographic Information System (GIS) was used \cite{QGIS_software}.

The hardware utilized was an NVIDIA GeForce Titan V100 GPU with 32 GB memory, as well as the NCCS Prism GPU Cluster(\url{https://www.nccs.nasa.gov/systems/ADAPT/Prism}). 

\section{Results}
\subsection{Fire Detection}

Table \ref{tab:fd_1} summarizes the performance of our SIT-FUSE-based segmentation approach, compared against the hand-labeled high certainty wildfire front areas, for wildfire fronts across all scenes in the test set, for each instrument tested. Table \ref{tab:fd_2} presents the same, but only for fires and geographic areas not seen during training. In all cases, our approach performs well. When visually compared, our approach tends to over-segment relative to the high-certainty fire pixels. However, the vast majority of the over-segmentation is over areas burning at lower heat, which are therefore less visibly associated with the fire front. This effect can be seen for the Williams Flats fire in Figure \ref{fig:master_wf} and for the Sheridan Fire in Figure \ref{fig:master_sheridan}.
\noindent
\belowrulesep = 0mm
\aboverulesep = 0mm
\begin{table}[ht!]
\centering
\caption {Summary of fire detection comparisons against the hand-labeled high-certainty wildland fire fronts over the full set of test scenes that contain fires. Total pixel count is the total number of pixels tested. SSIM was the metric used for comparisons in this study. For VIIRS and MODIS, the data from both platforms is used collectively.}
\begin{tabularx}{\linewidth}{YYY}
\hline
\textbf{Dataset} & \textbf{Total Pixel Count} &\textbf{SSIM} \\
\hline
MASTER& 12438023& \textbf{0.73}\\
\hline
eMAS& 162644645& \textbf{0.87}\\
\hline
AVIRIS-C& 177233951& \textbf{0.89}\\
\hline
GOES-17& 1016785& \textbf{0.82}\\
\hline
VIIRS& 4759386& \textbf{0.88}\\
\hline
MODIS& 3262896& \textbf{0.88}\\
\hline
\label{tab:fd_1}
\end{tabularx}
\end{table}
\vspace{1cm} 
\noindent
\belowrulesep = 0mm
\aboverulesep = 0mm
\begin{table}[ht!]
\centering
\caption {Summary of fire detection comparisons against the hand-labeled high-certainty wildland fire fronts over only the scenes that contain the Sheridan Fire and Horsefly Fire, unseen during training. Total pixel count is the total number of pixels tested. SSIM was the metric used for comparisons in this study. For VIIRS and MODIS, the data from both platforms is used collectively.}
\begin{tabularx}{\linewidth}{YYY}
\hline
\textbf{Dataset} & \textbf{Total Pixel Count} & \textbf{SSIM} \\
\hline
MASTER& 1708746& \textbf{0.89}\\
\hline
eMAS& 133762591& \textbf{0.88}\\
\hline
AVIRIS-C& 21073834& \textbf{0.93}\\
\hline
GOES-17& 2541975& \textbf{0.84}\\
\hline
VIIRS& 1189846& \textbf{0.88}\\
\hline
MODIS& 815724& \textbf{0.88}\\
\hline
\label{tab:fd_2}
\end{tabularx}
\end{table}
\end{paracol}

\begin{figure}[ht!]
\widefigure
\centering
    \subfloat{\includegraphics[width=0.7\textwidth, height=8cm]{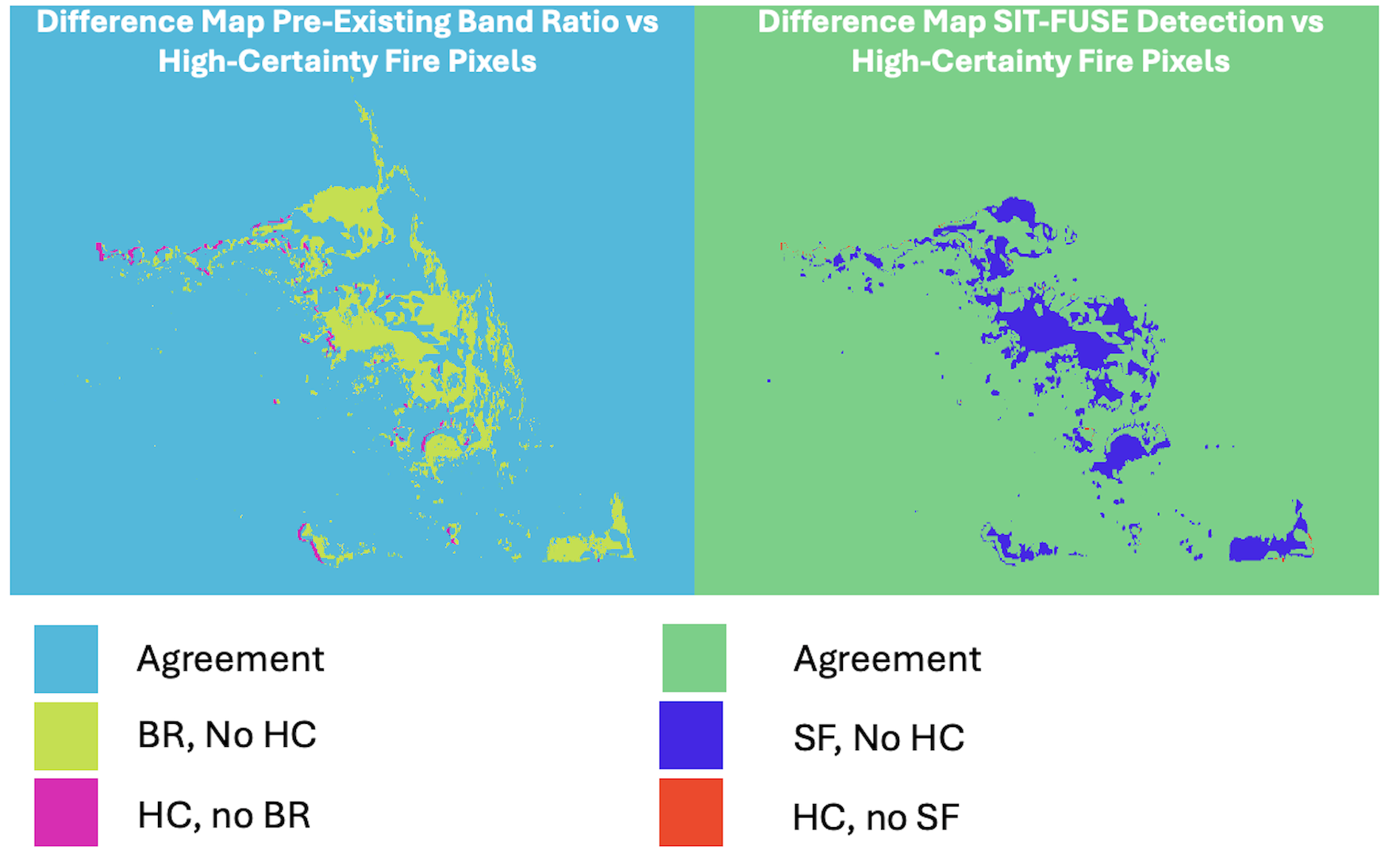}}
     \caption{A pair of difference maps between different detection outputs and the hand-labeled high-certainty fire pixels. The left panel is a difference map between the band-ratio-based detection (BR) and the high-certainty labels (HC). The right panel is a difference map between the SIT-FUSE-based detection (SF) and the high-certainty labels (HC).}
     \label{fig:master_diff_fire}
\end{figure}

\begin{figure}[ht!]
\widefigure
\centering
     \subfloat{\includegraphics[width=0.85\textwidth, height=7cm]{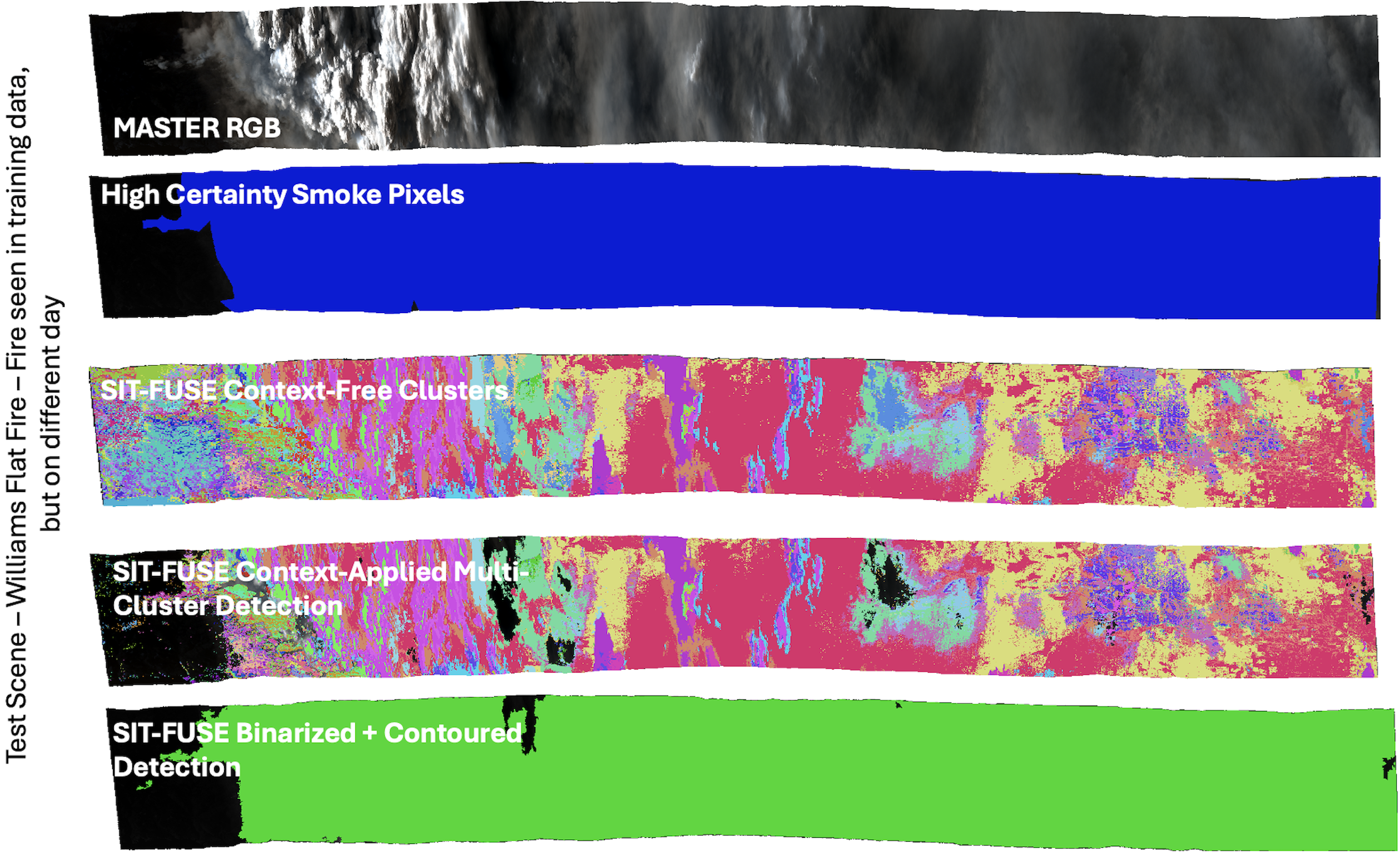}}
     \caption{The top left panel is an RGB image extracted from a MASTER scene over the Williams Flats fire. Following this panel, on the top row are the hand-labeled high-certainty fire pixels overlayed on the RGB image. The bottom row consists of the context-free segmentation map generated from SIT-FUSE for this scene, the context-applied multi-cluster detection - a subset of the full segmentation map with only the labels/clusters that correlate to smoke plumes, and the final binarized and contoured smoke mask.}
     \label{fig:master_smoke}
\end{figure}

\begin{figure}[ht!]
\widefigure
\centering
     \subfloat{\includegraphics[width=0.85\textwidth, height=9cm]{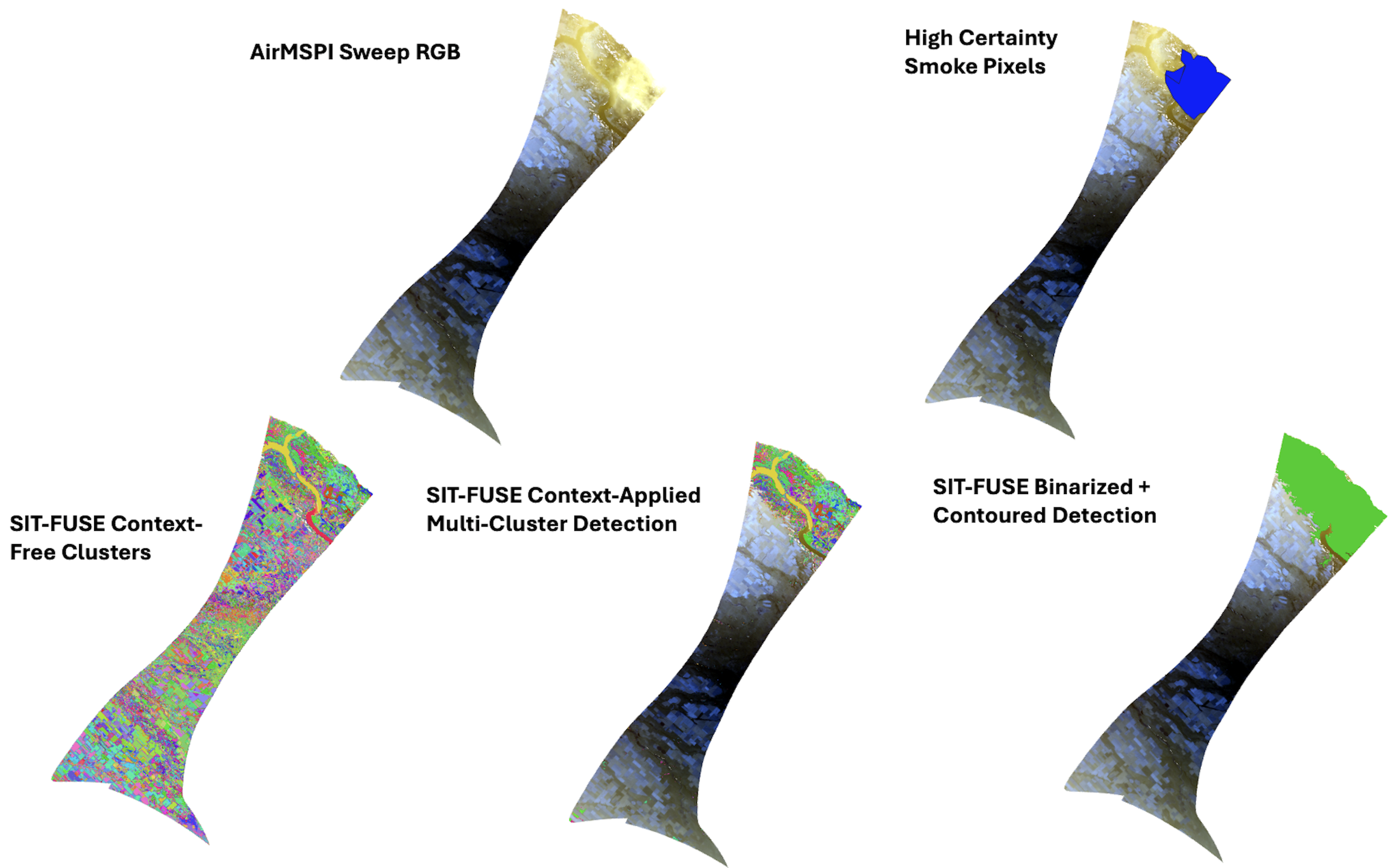}}
     \caption{The top left panel is an RGB image extracted from an AirMSPI scene over the Williams Flats fire. Following this panel, on the top row are the hand-labeled high-certainty fire pixels overlayed on the RGB image. The bottom row consists of the context-free segmentation map generated from SIT-FUSE for this scene, the context-applied multi-cluster detection - a subset of the full segmentation map with only the labels/clusters that correlate to smoke plumes, and the final binarized and contoured smoke mask.}
     \label{fig:mspi_smoke}
\end{figure}

\begin{paracol}{2}
\nolinenumbers
\switchcolumn

Table \ref{tab:fd_3} summarizes the same type of comparisons, this time against SIT-FUSE-based detections, and results from an experimental instrument-specific band-ratio-based detection methodology are compared against the high-certainty fire pixel label set for MASTER. For this comparison, we used all scenes in the test set that also had an associated band-ratio-based detection to compare against. The segmentation quality is quite comparable, with both techniques getting the same SSIM score, however, the difference maps in Figure \ref{fig:master_diff_fire} show a test-set wide trend of hot smoke also being picked up as part of the fire front in the band-ratio-based detection.

\noindent
\belowrulesep = 0mm
\aboverulesep = 0mm
\begin{table}[ht!]
\centering
\caption {Summary of fire detection comparisons against the hand-labeled high-certainty wildland fire fronts over only the scenes that contain the Sheridan Fire and Horsefly Fire, unseen during training. Total pixel count is the total number of pixels tested. SSIM was the metric used for comparisons in this study. }
\begin{tabularx}{\linewidth}{YYY}
\hline
\textbf{Dataset} & \textbf{Total Pixel Count} & \textbf{SSIM} \\
\hline
MASTER SIT-FUSE& 20389114& \textbf{0.8}\\
\hline
MASTER Pre-Existing& 20389114& \textbf{0.8}\\
\hline
GOES SIT-FUSE& 1016785& \textbf{0.82}\\
\hline
GOES Pre-Existing& 1016785& \textbf{0.72}\\
\hline
VIIRS SIT-FUSE& 4759386& \textbf{0.71}\\
\hline
VIIRS Pre-Existing& 4759386& \textbf{0.59}\\
\hline
\label{tab:fd_3}
\end{tabularx}
\end{table}

\subsection{Smoke Detection}

Table \ref{tab:sd_1} summarizes the performance of our SIT-FUSE-based segmentation approach, compared against the hand-labeled high certainty smoke plume pixels, for plumes across all scenes in the test set, for each instrument tested. Table \ref{tab:sd_2} presents the same specifically for smoke plumes in geographic areas not seen during training. In all cases, our approach performs well. When visually compared to the hand-labeled data, our approach tends to over-segment relative to the high-certainty smoke plume areas but performs well across the majority of the plume regions tested. The borders of the smoke plumes appear to provide uncertainty for both labelers and the SIT-FUSE-based automated detection. In some cases, these areas were not fully covered by our detections, whereas in other cases, the detections identified smoky areas in the scene that were not covered by the labels. Shadows also appear to confuse small numbers of cases. Figure \ref{fig:master_smoke} provides an example detection produced for a MASTER scene capturing the Williams Flats Fire, while Figure \ref{fig:mspi_smoke} shows the same example for an AirMSPI scene capturing the same fire.
\noindent
\belowrulesep = 0mm
\aboverulesep = 0mm
\begin{table}[ht!]
\centering
\caption {Summary of smoke detection comparisons against the hand-labeled high-certainty smoke plumes over the full set of test scenes that contain fires. Total pixel count is the total number of pixels tested. SSIM was the metric used for comparisons in this study.}
\begin{tabularx}{\linewidth}{YYY}
\hline
\textbf{Dataset} & \textbf{Total Pixel Count} &\textbf{SSIM}\\
\hline
MASTER& 12438023& \textbf{0.71}\\
\hline
eMAS& 162644645& \textbf{0.63}\\
\hline
AVIRIS-C& 173334623& \textbf{0.67}\\
\hline
AirMSPI Sweep& 132238386& \textbf{0.63}\\
\hline
PlanetScope Dove & 1353594326&\textbf{0.68}\\
\hline
GOES-17& 1016785& \textbf{0.62}\\
\hline
VIIRS& 4759386& \textbf{0.66}\\
\hline
\label{tab:sd_1}
\end{tabularx}
\end{table}
\noindent
\belowrulesep = 0mm
\aboverulesep = 0mm
\begin{table}[ht!]
\centering
\caption {Summary of smoke detection comparisons against the hand-labeled high-certainty fire areas over only the scenes that contain the Sheridan Fire and Horsefly Fire, unseen during training. Total pixel count is the total number of pixels tested. SSIM was the metric used for comparisons in this study. Planet data was not available over the Sheridan fire, so we cannot assess this.}
\begin{tabularx}{\linewidth}{YYY}
\hline
\textbf{Dataset} & \textbf{Total Pixel Count} &\textbf{SSIM} \\
\hline
MASTER& 1708746& \textbf{0.75}\\
\hline
eMAS& 133762591& \textbf{0.66}\\
\hline
AVIRIS-C& 21073834& \textbf{0.77}\\
\hline
AirMSPI Sweep& 63505132& \textbf{0.65}\\
\hline
PlanetScope Dove & N/A & \textbf{N/A}\\
\hline
GOES-17& 2541975& \textbf{0.65}\\
\hline
VIIRS& 1189846& \textbf{0.68}\\
\hline
\label{tab:sd_2}
\end{tabularx}
\end{table}
Table \ref{tab:sd_3} summarizes the same set of comparisons, this time evaluating SIT-FUSE-based detections against results from operational dark-target-based detection methodologies. Both approaches are compared against the high-certainty fire pixel label set for eMAS, GOES, and VIIRS. For these comparisons, we used the aerosol optical depth value from the dark target product of each of the instruments and applied a minimum threshold value of 0.2.
\noindent
\belowrulesep = 0mm
\aboverulesep = 0mm
\begin{table}[ht!]
\centering
\caption {Summary of smoke detection comparisons against the hand-labeled high-certainty smoke plumes over only the scenes unseen during training. Total pixel count is the total number of pixels tested. SSIM was the metric used for comparisons in this study. }
\begin{tabularx}{\linewidth}{YYY}
\hline
\textbf{Dataset} & \textbf{Total Pixel Count} & \textbf{SSIM} \\
\hline
eMAS SIT-FUSE& 162644645& \textbf{0.63}\\
\hline
eMAS Pre-Existing& 162644645& \textbf{0.37}\\
\hline
GOES SIT-FUSE& 1016785& \textbf{0.73}\\
\hline
GOES Pre-Existing& 1016785& \textbf{0.56}\\
\hline
VIIRS SIT-FUSE& 4759386& \textbf{0.66}\\
\hline
VIIRS Pre-Existing& 4759386& \textbf{0.54}\\
\hline
\label{tab:sd_3}
\end{tabularx}
\end{table}

\end{paracol}

\begin{figure}[th!]
\widefigure
\centering
    \subfloat{\includegraphics[width=0.7\textwidth, height=8cm]{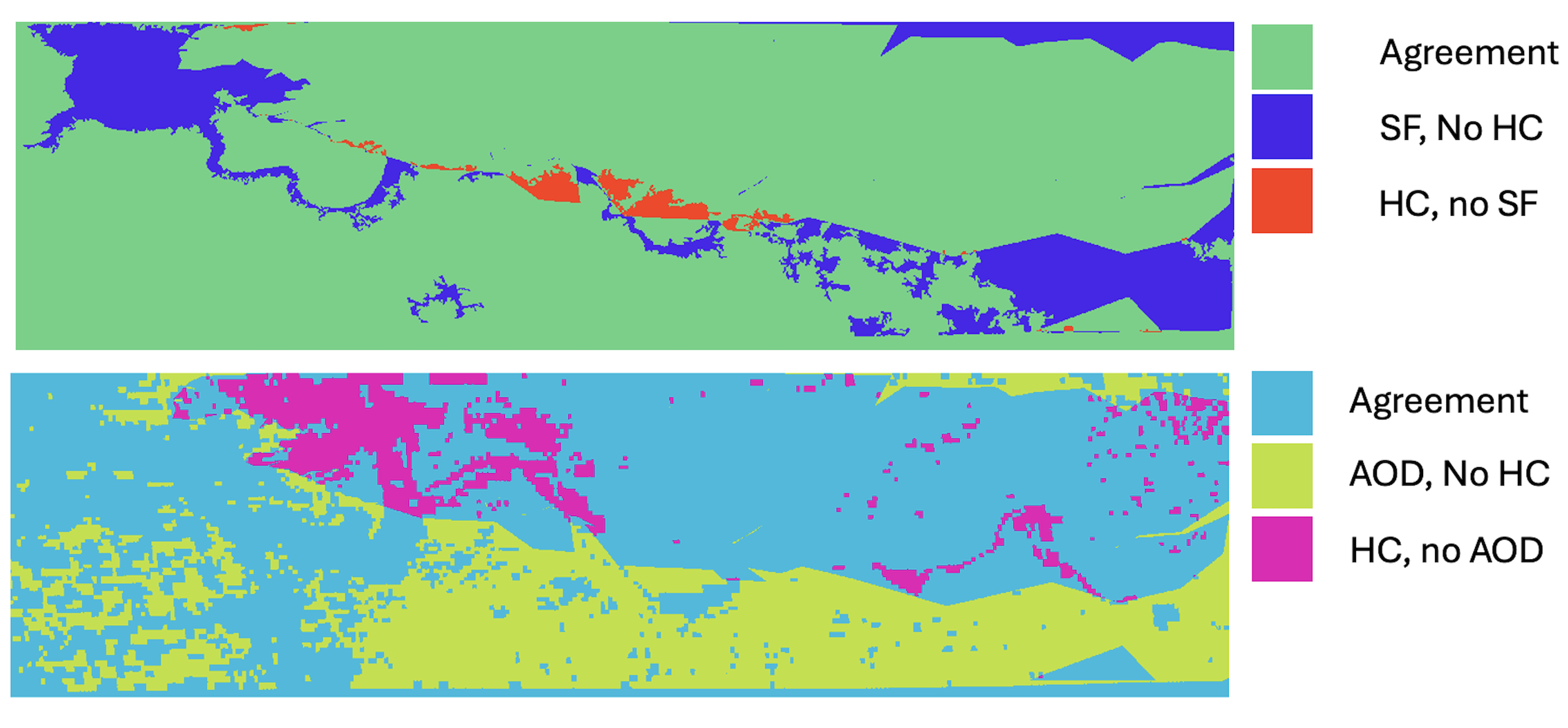}}
     \caption{A pair of difference maps between different detection outputs and the hand-labeled high-certainty smoke pixels. The top panel is a difference map between the SIT-FUSE-based detection (SF) and the high-certainty labels (HC). The bottom panel is the difference between the operational eMAS aerosol optical depth data, thresholded at 0.2 (AOD), and the high-certainty labels (HC).}
     \label{fig:emas_diff}
\end{figure}

\begin{figure}[ht!]
\widefigure
\centering
     \subfloat{\includegraphics[width=0.9\textwidth, height=8cm]{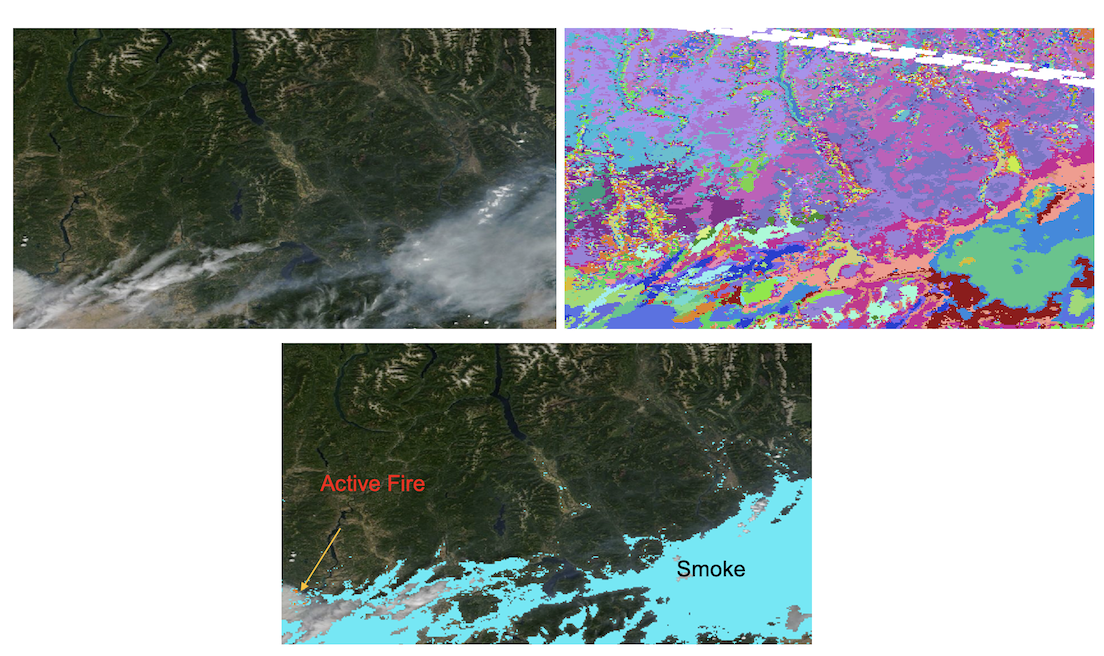}}
     \caption{Example wildfire front and smoke plume detection using Terra MODIS and 9-angle MISR scene over the Williams Flats fire on August 8, 2019, as input. The top left panel is a reference RGB generated from the MODIS data. The right panel is the context-free segmentation map generated from SIT-FUSE, and the bottom panel is the extracted and binarized \textcolor{blue}{smoke} and \textcolor{red}{fire} front detections from the SIT-FUSE outputs.}
     \label{fig:misr_modis}
\end{figure}

\begin{paracol}{2}
\nolinenumbers
\switchcolumn

\subsection{Multi-Sensor Fusion}

In previous work, we have demonstrated the utility of fusing multi-angle data from MISR with that of MODIS, both instruments onboard the same satellite platform, Terra. This included segmenting the William's Flats fire and Smoke plume in our evaluation. Figure \ref{fig:misr_modis} depicts segmentation, as well as fire and smoke segmentation over the Williams Flats fire, using MISR and MODIS data as input. However, combining MISR and MODIS data is limited by the restricted number of overpasses for a single platform and the narrower area of swath intersection, which reduces the number of cases available for evaluation compared to those in the single-instrument analyses described earlier. For polar-orbiting instruments, multiple platforms often provide an increased number of scenes for detection and evaluation. As we expand the spatiotemporal areas studied, we will add these fusion cases and quantify their benefits in a manner consistent with the previous evaluations.

\end{paracol}

\begin{figure}[ht!]
\widefigure
\centering
     \subfloat{\includegraphics[width=0.9\textwidth, height=10cm]{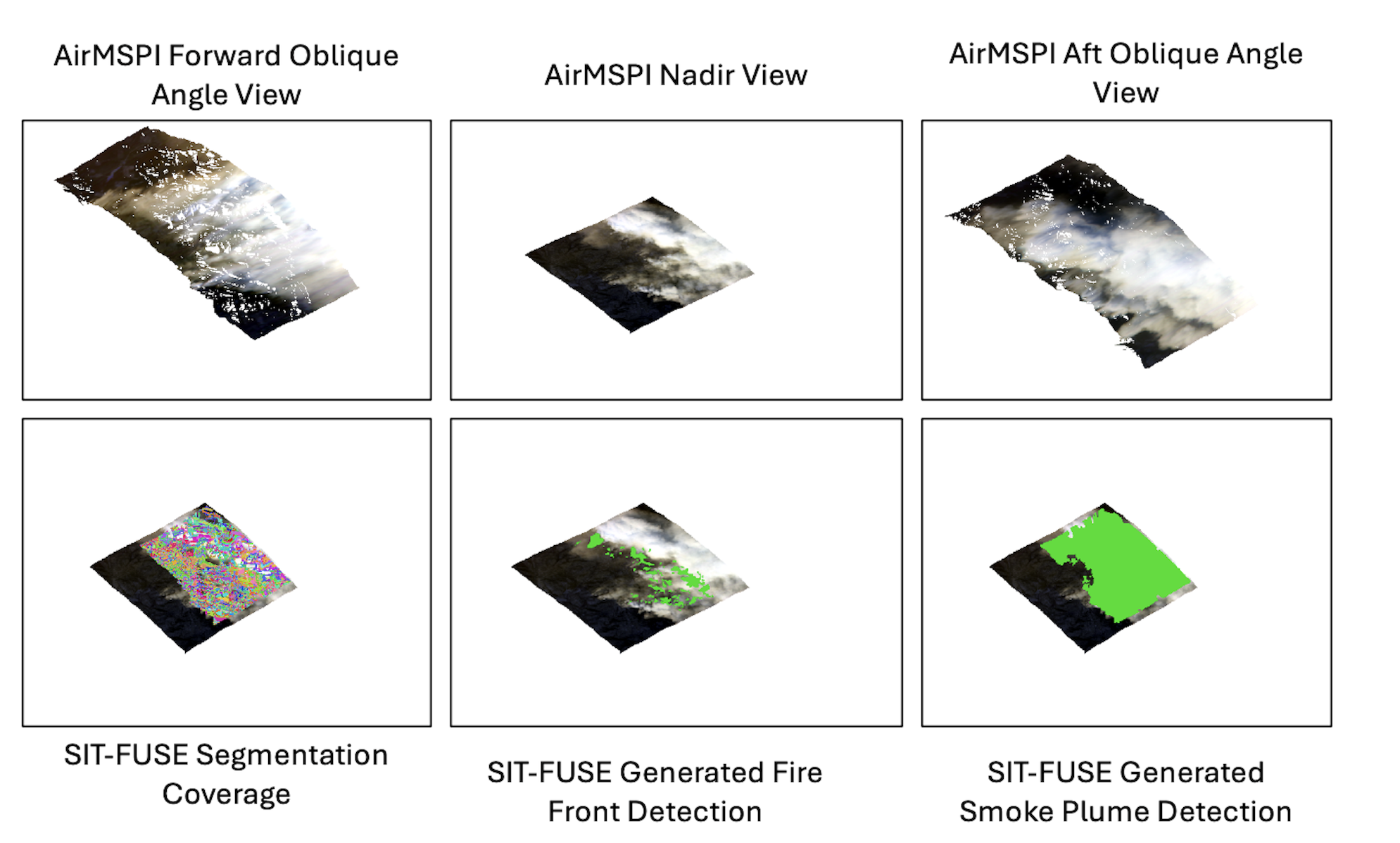}}
     \caption{Example wildfire front and smoke plume detection using multi-angle AirMSPI step-and-stare data. The panels on the top row are RGBs from the two most oblique angles, split by the RGB generated from the nadir data from a scene over the Williams Flats fire on August 06, 2019. The bottom row depicts the full coverage for the SIT-FUSE segmentation map, overlayed on the nadir RGB for reference, the extracted \textcolor{green}{fire front} detection, and the extracted \textcolor{green}{smoke plume} detection respectively.}
     \label{fig:mspi_sas}
\end{figure}

\begin{paracol}{2}
\nolinenumbers
\switchcolumn

Given our previous success with the fusion of polar-orbiting instrument data, we explored fusing airborne instrumentation data, both within the same platform and across platforms. For the instruments on the same platform, we examined AirMSPI and eMAS, both onboard the ER-2 aircraft, as an airborne equivalent of MISR and MODIS. As a first step, we tested AirMSPI's multi-angle data taken in step-and-stare mode collectively as a simple fusion test case. Unfortunately, the coverage of the data is very small and the number of scenes is relatively low. While there is spatial overlap between data captured at different viewing angles, data quality issues restrict its utility. There are areas marked as "bad/occluded data" that increased in frequency with more oblique angles. We tried to mitigate this by limiting the number of angles used, but it did not significantly improve, even after excluding data from four angles. Although we were able to generate smoke and fire detection examples, the number of scenes and pixels was insufficient for both model training and separate evaluation. Figure \ref{fig:mspi_sas} depicts an example scene from AirMSPI data, including the two most oblique angles and the nadir angle, as well as the minimal segmentation coverage we can get for SIT-FUSE when combining the multi-angle data, and the associated fire and smoke masks. 

For the AirMSPI + eMAS case, we also attempted to combine the sweep mode data with eMAS data. However, the overlap between the datasets was minimal. Future campaigns may take into account the need to fuse data in this manner and attempt to collect a greater number of spatiotemporally collocated scenes with multiple instruments. Again, as we expand our case studies and datasets over larger areas and campaigns, we are confident that we can quantitatively assess the benefits of data fusion between AirMSPI and eMAS for wildfire and smoke detection.

\end{paracol}

\begin{figure}[ht!]
\widefigure
\centering
     \subfloat{\includegraphics[width=0.9\textwidth, height=10cm]{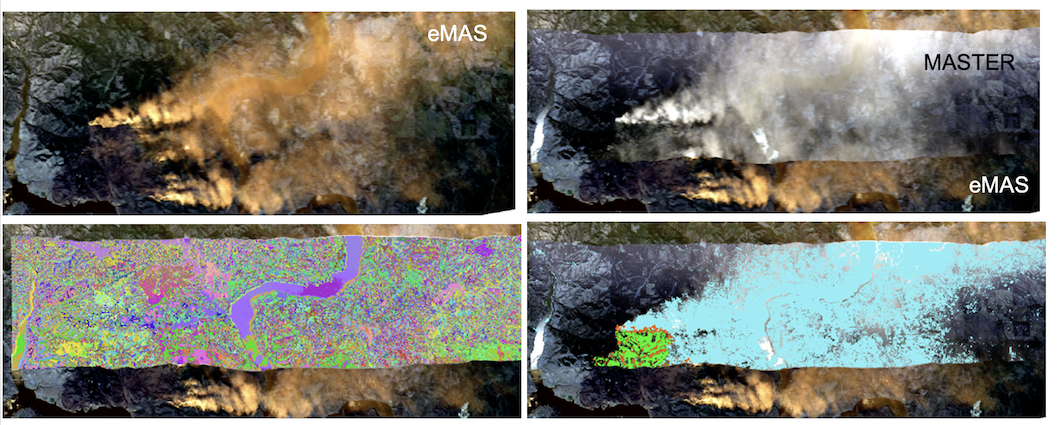}}
     \caption{Example wildfire front and smoke plume detection using MASTER and eMAS data as input. The top row depicts an RGB from an eMAS scene on August 06, 2019, and an RGB generated from the collocated MASTER scene overlayed on the eMAS RGB, both of which are used as input. The bottom row contains the context-free segmentation map overlayed on the area that contains both MASTER and eMAS data, and the extracted and binarized \textcolor{blue}{smoke} and \textcolor{red}{fire} front detections from the SIT-FUSE outputs.}
     \label{fig:master_emas}
\end{figure}

\begin{paracol}{2}
\nolinenumbers
\switchcolumn

For the case of cross-platform fusion, we looked at fusing eMAS and MASTER. Again the number of scenes is limited, and only over the Williams Flats fire area, but the initial examples demonstrate potential for this capability, and again future campaign planning and use of additional pre-existing campaign datasets will allow us to further validate these fusion cases. Figure \ref{fig:master_emas} depicts an example of one of the fusion scenes.

\end{paracol}

\begin{figure}[ht!]
\widefigure
\centering
     \subfloat{\includegraphics[width=0.9\textwidth, height=12cm]{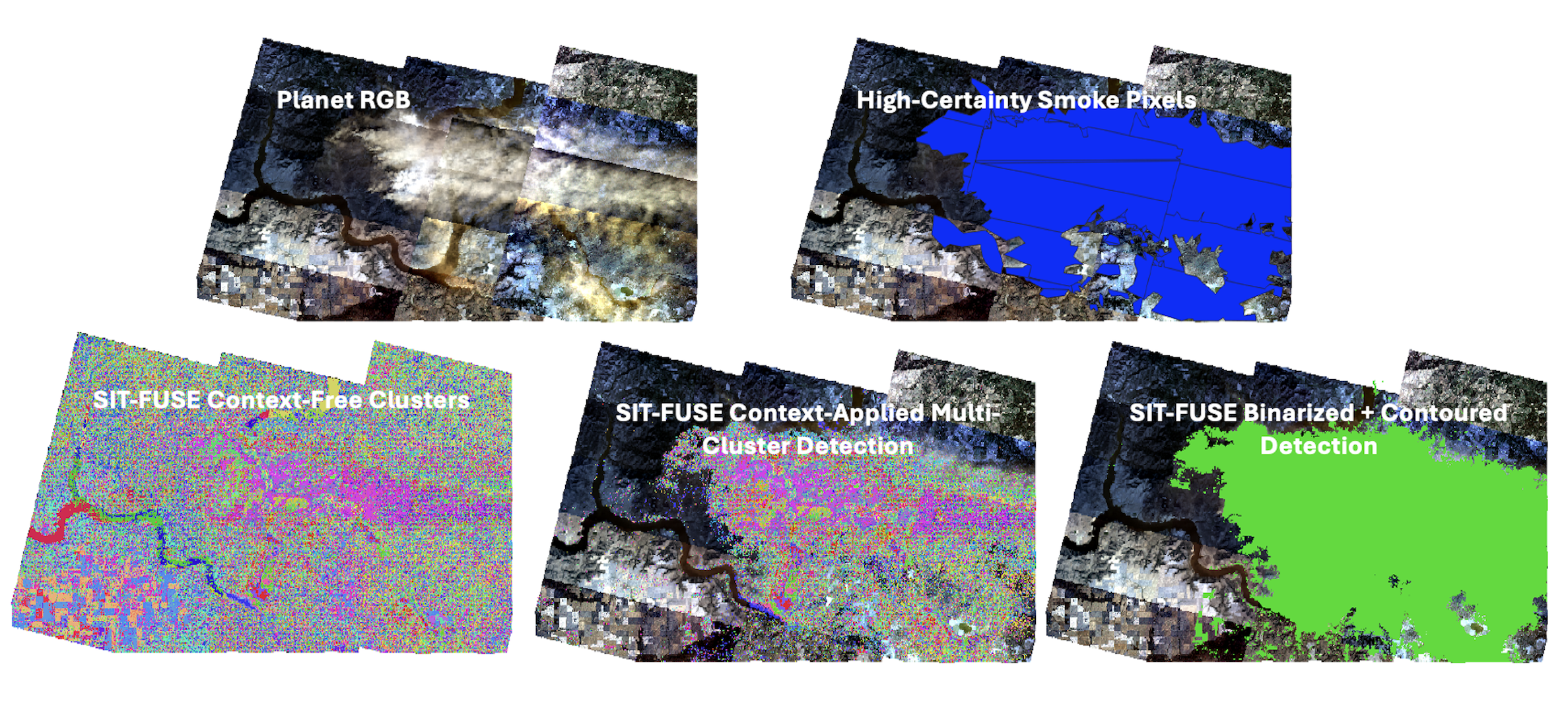}}
     \caption{An example of using Planet data to do smoke detection. The top row depicts a mosaic of RGBs generated from the input scenes and the high-certainty smoke pixels overlayed on top. The bottom row contains the context-free segmentation map generated from SIT-FUSE, the subset labels/clusters that are associated with smoke, and the final panel contains the contoured and binarized smoke plume detected over the entire mosaic. Image \textsuperscript{\textcopyright} 2019 Planet Labs PBC}
     \label{fig:planet_mosaic}
\end{figure}

\begin{paracol}{2}
\nolinenumbers
\switchcolumn

\subsection{Commercial Remote Sensing Products}

As a part of this effort, we evaluated the utility of PlanetScope data for wildfire and smoke detection and monitoring. We found that the older PlanetScope SuperDove imagers available during the FIREX-AQ 2019 campaign were not suited for identifying fire fronts across the entire test set. However, they were effective at detecting smoke within the scenes evaluated. Figure \ref{fig:planet_mosaic} depicts smoke detection within a mosaic of SuperDove scenes over the Williams Flats fire on August 08, 2019, and Table \ref{tab:sd_1} contains quantitative evaluation results. The improved temporal resolution of smoke detections at high spatial resolution will be helpful both for air quality research and automated smoke tracking. Also, as Planet's instrument offerings continue to diversify, with new SuperDove instruments having a higher spectral resolution, hyperspectral instrumentation being launched, and other companies continuing to launch various other Earth-orbiting instrumentation, it is critical that we look at the tradeoffs between paying for data from commercial entities and the scientific value added by such data. In this study, we analyzed 83 additional scenes over the Williams Flats area over four days. Per the NASA / Planet CSDSA agreement, we are not able to release the input SuperDove data, alongside all of the other data we are releasing. However, we will release the generated output and all other associated data.

\end{paracol}

\begin{paracol}{2}
\nolinenumbers
\switchcolumn

\end{paracol}

%

\begin{paracol}{2}
\nolinenumbers
\switchcolumn
\section{Discussion and Current/Future Work}
Overall, SIT-FUSE effectively identifies and segments wildland fire fronts and smoke plumes, and performs better against high-certainty smoke and fire label sets, when compared against other operational and experimental approaches, as seen in Tables \ref{tab:fd_3} and \ref{tab:sd_3}. As expected, the higher the resolution of the instrument, the more detailed segmentation can occur. However, the different resolutions are valuable not only as standalone datasets but also for tracking these objects across multiple datasets. While further large-scale validation is needed, our tests demonstrate the capability to increase the temporal resolution of firefront products. We do this by increasing the number of instruments available for fire and smoke segmentation and developing brand-new firefront and smoke plume products for many of the instruments tested. This indicates significant potential for both product generation and utilization of this technique for segmentation and instance tracking. This not only also allows for dynamic instance tracking across scenes with the same input set, but we believe by harnessing style transfer capabilities, we can also look into instance tracking across multi-sensor scenes from disparate input datasets. We hope to transition this technology into a piece of future operational campaign support or product generation systems.

In terms of feature interpretability and selection, methods such as SHAP analysis and other explainability methods can be applied to better understand feature importance and refine the input to focus on spectral bands most effective for identifying smoke and/or fire. Given the current performance and the success with datasets where there was no pre-existing operational fire or smoke detection methodology, solutions like SIT-FUSE can be integrated into new or existing instrumentation data processing pipelines. By doing so, this approach could replace or augment instrument-specific retrieval algorithms, which may be extremely costly to develop. SIT-FUSE's segmentation capabilities offer additional benefits: the decrease in data volume processed for downstream fire or smoke-specific retrievals. By isolating the detected objects, only relevant pixels need to be processed through a downstream retrieval, thereby optimizing the pipeline.

We have built a framework within SIT-FUSE that is adaptable to various kinds of encoders and we aim to be able to leverage this to analyze representative capabilities of different model types, complexities, and training paradigms. With the continued influx of new architectures and large Earth Observation Foundation Models, it is important to understand these models provide quality representations (or poor ones) under different conditions, problem sets, and input datasets \cite{pangaea}. Analyses of downstream task performance is a crucial piece, but not the entire solution. More robust ways to evaluate representative capabilities are emerging in around large language models (LLMs), and much of this can be ported to computer vision, and specifically deep learning for Earth Observations \cite{kernels}. With the flexible framework of SIT-FUSE we are working towards providing initial pathways towards tackling some of these open problems.

Lastly, we are working to leverage SIT-FUSE to make an impact within the area of analysis and scientific understanding - in this case correlated to wildfires and smoke plumes. There a built-in co-discovery facilitation mechanism, by way of the heirarchichal context-free segmentation products. By using the model-derived separations of various areas, novelty and "interesting" samples can more easily be grouped and investigated. This can be even further coupled with more detailed analyses of the embedding spaces relative to the context-free segmentations \cite{encoder_comp}. To enhance exploration even further models trained for co-exploration of data using open-ended algorithms can be leveraged to more quickly sift through the volumes of data and highlight interesting, new, and anomoulous samples \cite{OMNI, ai_sci}.

\authorcontributions{Conceptualization, Nicholas LaHaye, Kyongsik Yun, Michael Garay, and Erik Linstead; Data curation, Nicholas LaHaye, Anastasija Easley, Huikyo Lee, Michael Garay and Olga Kalashnikova; Formal analysis, Nicholas LaHaye and Anastasija Easley; Funding acquisition, Nicholas LaHaye, Kyongsik Yun, Huikyo Lee, Michael Garay, and Olga Kalashnikova; Investigation, Nicholas LaHaye, Kyongsik Yun, Michael Garay, Olga Kalashnikova and Erik Linstead; Methodology, Nicholas LaHaye, Kyongsik Yun, Huikyo Lee, and Erik Linstead; Project administration, Nicholas LaHaye, Software, Nicholas LaHaye, Kyongsik Yun, Huikyo Lee, and Erik Linstead; Supervision, Nicholas LaHaye, Michael Garay, Olga Kalashnikova and Erik Linstead; Validation, Nicholas LaHaye, Kyongsik Yun, Huikyo Lee, Michael Garay, Olga Kalashnikova and Erik Linstead; Visualization, Nicholas LaHaye, Kyongsik Yun, Huikyo Lee, and Erik Linstead; Writing – original draft, Nicholas LaHaye; Writing – review \& editing, Nicholas LaHaye, Kyongsik Yun, Huikyo Lee, Michael Garay, Olga Kalashnikova and Erik Linstead.}

\funding{This research was funded by the NASA ROSES Commercial Smallsat Data Scientific Analysis program (NNH22ZDA001N-CSDSA) as well as the Jet Propulsion Laboratory's Data Science Working Group. Computing resources were leveraged at both the NASA Center for Climate Simulation (NCCS) and the Machine Learning and Affiliated Technologies (MLAT) Lab in the Fowler School of Engineering at Chapman University.}

\dataavailability{The data has been published and is freely available on Zenodo \cite{project_data}. The model weights and associated configuration files are also publicly available on HuggingFace \cite{project_model_weights}. Lastly, the code has been tagged at the time of this paper submission and is publicly availabel on GitHub \cite{sit_fuse_software}. The Planet radiance data cannot be released publicly due to the terms of the NASA Commercial Smallsat Data Acquisition agreement, but all products generated and models trained have been shared, and all other input data is freely available in the afforementioned Zenodo data store.} 

\acknowledgments{The authors would like to thank NASA, JPL, NCCS, MLAT Lab., The Spatial Informatics Group, LLC., and the Schmid College of Science and Technology, Chapman University, for supporting this research. The authors would also like to thank Phil Dennison, from the Geography Department, at the University of Utah, as well as NASA for providing the data, without which this research would not have been possible. Finally, the authors would like to thank the anonymous reviewers for taking the time to read this paper and provide valuable feedback.}

\conflictsofinterest{The funders had no role in the design of the study; in the collection, analyses, or interpretation of data; in the writing of the manuscript, or in the decision to publish the results.} 

\end{paracol}

\reftitle{References}


\bibliography{FIREX.bib}




\end{document}